%% file: main.tex
\documentclass{article}

    \PassOptionsToPackage{numbers, compress,sort}{natbib}

\usepackage{arxiv}

\usepackage{natbib}[numbers,sort,compress]

\usepackage[utf8]{inputenc} 
\usepackage[T1]{fontenc}    
\usepackage{hyperref}       
\usepackage{url}            
\usepackage{booktabs}       
\usepackage{amsfonts}       
\usepackage{nicefrac}       
\usepackage{microtype}      
\usepackage{xcolor}         
\input{Utils/mysymbols}
\input{Utils/math_utils}

\everydisplay{\small}

\title{
{\ours}: Towards Lossless Speedup for Reasoning Training through Edge-Preserving CoT Condensation
}

\author{Jinghan Jia$^{\dag}$ ~~Hadi Reisizadeh$^{\P}$ ~~Chongyu Fan$^{\dag}$ ~~Nathalie Baracaldo$^{\S}$ ~~Mingyi Hong$^{\P,\ddag}$ ~~Sijia Liu$^{\dag,\S}$\\
  $^\dag$Michigan State University\\
  $^\ddag$Amazon\\
  $^\P$University of Minnesota\\
  $^\S$IBM Research\\
}
\date{}

\begin{document}

\maketitle

\input{sections/abstract}
\input{sections/intro}
\input{sections/related_work}
\input{sections/preliminary_SL}
\input{sections/method_SL}
\input{sections/experiments}

\input{sections/conclusion}
{{
\bibliographystyle{unsrtnat}
\bibliography{ref/reasoning}
}}

\clearpage
\newpage
\input{sections/appendix}

\end{document}

%% file: Utils/mysymbols.tex
\usepackage{wrapfig}
\usepackage{multirow,mathtools}
\usepackage{pifont}
\usepackage{color, colortbl}

\usepackage{blindtext}
\usepackage{lipsum}

\usepackage{multirow}
\usepackage{makecell}
\usepackage{graphicx}
\usepackage{listings}

\usepackage{bbm}

\usepackage [english]{babel}
\usepackage [autostyle, english = american]{csquotes}
\usepackage[nottoc]{tocbibind}

\usepackage{pifont}
\usepackage{url}
\usepackage[most]{tcolorbox}

\usepackage{lipsum}
\usepackage{xcolor}
\usepackage{wrapfig}
\usepackage{booktabs}

\usepackage{bbold}

\usepackage[most]{tcolorbox}
\usepackage{pifont}
\usepackage{multirow}
\usepackage{booktabs}
\usepackage{colortbl}

\definecolor{Gray}{gray}{0.93}
\definecolor{Orange}{rgb}{1,0.5,0}
\definecolor{DGray}{gray}{0.83}
\definecolor{modelrowcolor}{RGB}{204,229,255}
\definecolor{darkergreen}{RGB}{1, 50, 32}
\newcommand{\SL}[1]{\textcolor{purple}{SL: #1}}
\newcommand{\JH}[1]{\textcolor{blue}{JH: #1}}

\usepackage{color, colortbl}

\newcommand{\blue}{\color{blue}}

\usepackage{pifont}

\usepackage{color, colortbl}
\definecolor{Gray}{gray}{0.93}
\definecolor{Orange}{rgb}{1,0.5,0}
\definecolor{Green}{rgb}{0,0.80,0}
\definecolor{Blue}{rgb}{0,0,0.92}
\definecolor{Red}{rgb}{0.90,0,0}
\definecolor{DGray}{gray}{0.83}
\definecolor{LightCyan}{rgb}{0.88,1,1}

\definecolor{bluegray}{rgb}{0.4, 0.6, 0.8}
\definecolor{ceruleanblue}{rgb}{0.16, 0.32, 0.75}

\newcommand{\ours}{{{EPiC}}}
\newcommand{\HoC}{{{HoC}}}
\newcommand{\ToC}{{{ToC}}}
\newcommand{\MoC}{{{MoC}}}

%% file: Utils/math_utils.tex
\usepackage{amsmath,amsfonts,bm}

\def\eqref#1{(\ref{#1})}

\def\1{\bm{1}}

\DeclareMathAlphabet{\mathsfit}{\encodingdefault}{\sfdefault}{m}{sl}
\SetMathAlphabet{\mathsfit}{bold}{\encodingdefault}{\sfdefault}{bx}{n}

\DeclareMathOperator*{\minimize}{\text{minimize}}

\newcommand{\btheta}{{\boldsymbol{\theta}}}

\newcommand{\bfr}{\boldsymbol{r}}

\newcommand{\bfa}{\boldsymbol{\alpha}}

%% file: sections/abstract.tex
\begin{abstract}

Large language models (LLMs) have shown remarkable reasoning capabilities when trained with chain-of-thought (CoT) supervision. However, the long and verbose CoT traces, especially those distilled from large reasoning models (LRMs) such as DeepSeek-R1, significantly increase training costs during the distillation process, where a non-reasoning base model is taught to replicate the reasoning behavior of an LRM.
In this work, we study the problem of \textit{CoT condensation} for resource-efficient reasoning training, aimed at pruning intermediate reasoning steps (\textit{i.e.}, thoughts) in CoT traces, enabling supervised model training on length-reduced CoT data while preserving both answer accuracy and the model's ability to generate coherent reasoning. 
Our rationale is that CoT traces typically follow a three-stage structure: problem understanding, exploration, and solution convergence. Through empirical analysis, we find that retaining the structure of the reasoning trace, especially the early stage of problem understanding (rich in reflective cues) and the final stage of solution convergence (which closely relates to the final answer), is sufficient to achieve \textit{lossless} reasoning supervision. To this end, we propose an \underline{E}dge-\underline{P}reserv\underline{i}ng  \underline{C}ondensation method, \textbf{\ours}, which selectively retains only the initial and final segments of each CoT trace while discarding the middle portion. This design draws an analogy to preserving the ``edge'' of a reasoning trajectory, capturing both the initial problem framing and the final answer synthesis, to maintain logical continuity. 
Experiments across multiple model families (Qwen and LLaMA) and benchmarks show that {\ours} reduces training time by over 34\% while achieving lossless reasoning accuracy on \textsc{Math500}, comparable to full CoT supervision. To the best of our knowledge, this is the first study to explore thought-level CoT condensation for efficient reasoning model distillation. Codes are available at \,\url{https://github.com/OPTML-Group/EPiC}. 
\end{abstract}

%% file: sections/intro.tex
\section{Introduction}
\label{sec: intro}

Large language models (LLMs) have demonstrated strong performance on complex reasoning tasks, especially when trained with chain-of-thought (CoT) supervision~\cite{wei2022chain, lightman2023let, guo2025deepseek}. CoT training encourages models to generate step-by-step intermediate reasoning before producing a final answer, enhancing both interpretability and task performance in domains such as mathematics \cite{openai2024openaio1card} and science \cite{rein2024gpqa}. 
More recently, large reasoning models (LRMs) such as DeepSeek-R1~\cite{guo2025deepseek}, OpenAI-O1~\cite{openai2024openaio1card}, and Kimi~\cite{team2025kimi} have pushed this paradigm further by generating rich CoT traces infused with self-reflection, verification, and backtracking, \textit{e.g.}, acquired via reinforcement learning. These LRMs have enabled a new training pipeline: Their reasoning ability can be distilled into smaller LLMs via supervised fine-tuning (SFT) on LRM-generated CoT data~\cite{guo2025deepseek, openr1, openthoughts, muennighoff2025s1, ye2025limo}.
{Throughout this paper, we refer to training (non-reasoning) LLMs with CoT supervision (for reasoning enhancement) as \textit{reasoning training}.}

However, despite their quality, LRM-generated CoT traces are often excessively verbose and suffer from \textit{overthinking}, a tendency to include repetitive or speculative reasoning steps that inflate sequence length without improving final answer accuracy~\cite{chen2024not, wang2025thoughts}. This verbosity leads to two key issues: (1) high computational cost during SFT, and (2) reduced supervision quality due to noise, particularly in the middle of the trace where speculative exploration dominates.
These observations raise a central question: \textit{Are all reasoning steps equally important for reasoning training?}

\begin{figure}[t]
\centering
\includegraphics[width=0.90\textwidth]{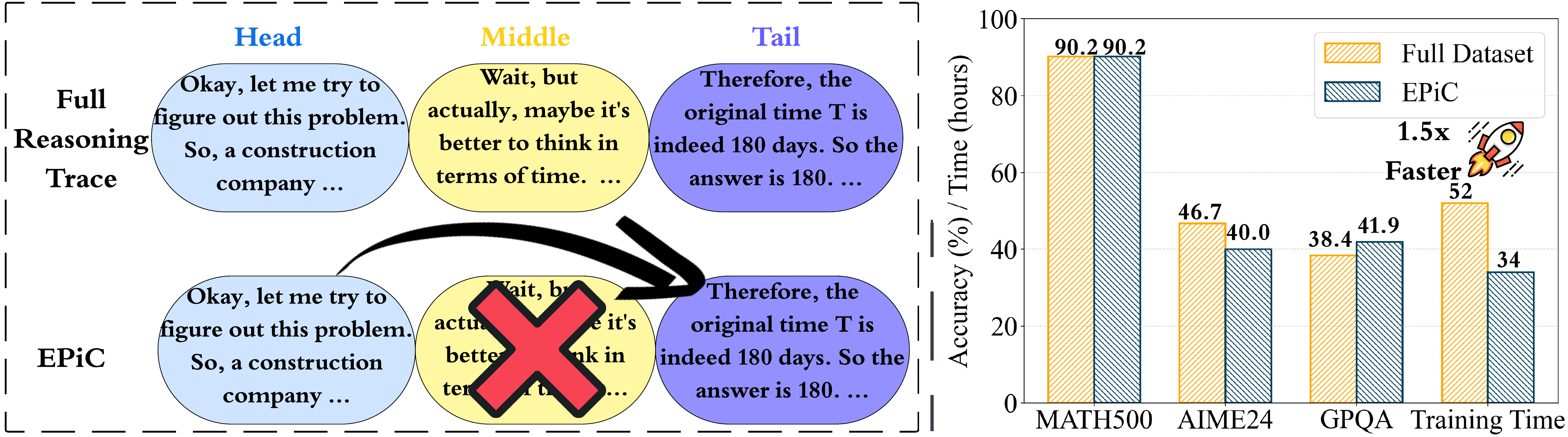} 
  \caption{\footnotesize{{Overview of {\ours}.} 
\textbf{Left:} {\ours} removes the middle portion of CoT while preserving the head (problem understanding) and tail (solution convergence). {\ours} applies to training data in OpenR1Math. 
\textbf{Right:} Performance and training time comparison between {\ours} and full CoT training based on \textsc{Qwen2.5-Math-7B-Instruct}. At 50\% condensation ratio, {\ours} achieves competitive accuracy with {1.5× faster} training.
}
  }
  \vspace*{-3mm}
  \label{fig:teaser_fig}
\end{figure}
In this work, we propose \underline{E}dge-\underline{P}reserv\underline{i}ng \underline{C}ondensation ({\ours}),  a simple yet effective thought-level pruning method that retains the head and tail segments of each CoT trace, corresponding to problem understanding and solution convergence, while removing only the middle portion of the reasoning trajectory.
As illustrated in \textbf{Figure\,\ref{fig:teaser_fig} (left)}, {\ours} targets the overgenerated middle stage, preserving the structural and semantic integrity of the reasoning process. As shown in \textbf{Figure\,\ref{fig:teaser_fig} (right)}, {\ours} enables models to achieve competitive reasoning accuracy while reducing training time by \textit{1.5×} compared to full-trace fine-tuning.


To better understand which segments of a CoT trace are most critical for reasoning supervision, we conduct a mutual information (MI) analysis between individual segments and the full reasoning trace. We find that the portions selected by {\ours} consistently exhibit the highest MI with the complete trace, supporting our empirical finding that the middle segment is the least informative (and often the noisiest) part of the reasoning trajectory. These findings motivate {\ours} as a principled and efficient strategy for CoT data-based reasoning training.
Our \textbf{contributions} are summarized as follows:


$\bullet$ We introduce the first framework {\ours} to perform thought-level condensation during \textit{training}, enabling efficient reasoning distillation by pruning uninformative steps within CoT traces.


$\bullet$ {We provide a series of analyses, including thought landscape visualization, mutual information analysis, and CoT perturbation studies, to quantify informativeness across CoT segments and validate the effectiveness of {\ours}.}


$\bullet$ 
{We conduct extensive experiments across two CoT training datasets (OpenR1Math and GeneralThought), three reasoning benchmarks (MATH500, AIME24, and GPQA-Diamond), and multiple non-reasoning model initializations (\textsc{Qwen2.5-Math-7B-Instruct}, \textsc{Qwen2.5-7B-Instruct}, and \textsc{LLaMA3.1-8B-Instruct}). Our results show that {\ours} consistently achieves final answer accuracy comparable or better than full CoT supervision, while significantly reducing training time.
}

%% file: sections/related_work.tex
\vspace*{-1mm}
\section{Related Work}
\vspace*{-1mm}

\paragraph{Model distillation for LRMs.}
CoT prompting has been shown to significantly improve the reasoning capabilities of large language models (LLMs) \citep{wei2022chain}, motivating a line of work that seeks to build LRMs through CoT-style data construction \citep{zhou2022teaching,shridhar2023distilling,fu2023specializing}. With the emergence of strong LRMs such as \textsc{OpenAI-O1} \citep{openai2024openaio1card}, \textsc{DeepSeek-R1} \citep{guo2025deepseek}, and \textsc{Kimi-1.5} \citep{team2025kimi}, which can autonomously generate long and structured CoT traces, including self-reflection, verification, and backtracking, researchers have increasingly focused on distilling such behaviors into smaller models. \citet{guo2025deepseek} was among the first to demonstrate that the reasoning capabilities of LRMs can be effectively transferred to smaller models through SFT. Building on this insight, numerous works have explored distillation from LRM-generated CoT data to improve reasoning performance in smaller LLMs \citep{openthoughts,muennighoff2025s1,ye2025limo,openr1,generalthought195k,skyt12025,bespokestratos,badri2025r1,li2025small,xu2025redstar,ji2025first}. These works can be broadly grouped into two categories: (1) distillation via high-quality, long-form CoT traces generated from LRMs \citep{openthoughts,muennighoff2025s1,ye2025limo,openr1,generalthought195k,skyt12025,bespokestratos,li2025small,xu2025redstar}; and (2) alignment-based approaches that directly supervise logits \citep{badri2025r1}. A complementary direction, explored by \citet{ji2025first}, proposes a hybrid strategy that uses truncated CoT prefixes with a small portion of full traces for efficient distillation. While prior work has successfully leveraged LRM-generated traces for performance improvement, only a few efforts \citep{muennighoff2025s1,ye2025limo} have addressed the efficiency bottlenecks in CoT distillation. These works focus on reducing the number of training examples while ensuring trace diversity and quality. In contrast, our work preserves all examples but reducing the length of each trace through structured thought-level pruning. Specifically, we study how to remove uninformative reasoning steps within each sample to reduce training cost, without compromising the effectiveness of the distilled model.

\paragraph{Scaling test-time reasoning and the challenge of overlength generation.} Increasing test-time computation has consistently improved model performance on complex reasoning tasks such as mathematical problem solving and code generation~\citep{wei2023chainofthoughtpromptingelicitsreasoning, wu2024inferencescalinglawsempirical, deepseekai2025deepseekr1incentivizingreasoningcapability, snell2024scalingllmtesttimecompute}. These gains often come from generating longer reasoning traces or sampling diverse reasoning paths~\citep{openai2024openaio1card, wu2024inferencescalinglawsempirical}. Recent methods include parallel path sampling~\citep{wang2023selfconsistencyimproveschainthought, aggarwal2023let, brown2024large}, tree-based search~\citep{yao2023treethoughtsdeliberateproblem, xin2024deepseekproverv15harnessingproofassistant}, and iterative refinement~\citep{welleck2023generating, madaan2023selfrefineiterativerefinementselffeedback, welleck2024decodingmetagenerationinferencetimealgorithms}. Additionally, \citet{muennighoff2025s1} proposed enhancing the use of reflection tokens at inference time, and others~\citep{snell2024scalingllmtesttimecompute, liu2025can} showed that scaling test-time computation can rival or exceed model size increases. However, these strategies often induce overthinking, verbose, repetitive outputs that slow inference and may reduce quality~\citep{chen2024not, wang2025thoughts}. This is especially common in LRMs, which tend to generate redundant reasoning steps and excessive self-reflection. To mitigate this, several methods promote concise, efficient reasoning: \citet{team2025kimi}, \citet{aggarwal2025l1}, and \citet{luo2025o1} introduce length-regularized RL; \citet{xia2025tokenskip} apply SFT with truncated or token-pruned inputs; \citet{wang2025thoughts} penalize reflection token usage; and \citet{zhang2025lightthinker} compress thoughts via token projection for faster decoding. While these approaches primarily target inference-time efficiency, our work addresses the complementary challenge of improving training-time efficiency. Specifically, we investigate how to condense reasoning trajectories during supervised fine-tuning, enabling small models to learn LRM-style reasoning with reduced computational cost.

\paragraph{Dataset pruning for efficient training.}  To reduce training costs, data pruning has been widely studied in discriminative settings like image classification, where redundant samples are removed~\citep{kothawade2021similar, killamsetty2021grad, lee2021deduplicating, azeemi2022dataset}. Importance score for each sample is estimated using geometry-based~\citep{agarwal2020contextual}, uncertainty-based~\citep{coleman2019selection}, margin-based~\citep{park2022active}, gradient-based~\citep{mirzasoleiman2020coresets}, forgetting-based~\citep{toneva2018empirical}, and training-dynamics-based methods~\citep{paul2021deep}, with learned pruners also explored~\citep{huang2023fewer}. These approaches have recently been adapted for LLM instruction tuning~\citep{zhangstaff, xia2024less}, and \citet{zhou2023lima} showed that strong performance can be achieved with just 1,000 high-quality examples. However, pruning for reasoning training remains underexplored. \citet{ye2025limo} leveraged a small set of curated CoT traces, but no prior work investigates pruning at the level of individual reasoning steps. In contrast, we introduce \textit{thought-level condensation}, a fine-grained strategy that prunes within examples rather than across them.

%% file: sections/preliminary_SL.tex
\section{Condensed CoT for Efficient Reasoning Training: Motivation and Problem}
\label{sec: pre}

In this section, we first reviews CoT-based reasoning training and its standard setup, then highlight the trade-off between efficiency and accuracy identified in prior work. Motivated by this tension, we investigate the potential of \textit{thought selection} and formally introduce the problem of \textit{CoT condensation}, which seeks to accelerate reasoning training without compromising reasoning performance.


\paragraph{Reasoning enhancement via supervised fine-tuning (SFT) on CoT data.} Training LLMs to reason step by step, rather than directly predicting final answers, using CoT supervision has shown significant promise \citep{guo2025deepseek, jaech2024openai, xu2025redstar,min2024imitate}. This reasoning training approach has emerged as an effective model distillation technique, enabling smaller, non-reasoning LLMs to acquire reasoning abilities by fine-tuning on long CoT traces generated by larger models.

This work focuses on improving the efficiency of reasoning training with CoT supervision, aiming to achieve faster training while maintaining or even enhancing reasoning capabilities, as measured by final answer accuracy on complex problems (\textit{e.g.}, mathematics) and the ability to generate coherent reasoning traces (\textit{e.g.}, reflected in the length of the distilled model outputs).

To be more concrete, let $\mathcal{D} = \{(\mathbf{x}, \mathbf{r}, \mathbf{y})\}$ denote a CoT-style training dataset, where $\mathbf{x}$ is the input question, $\mathbf{r} = [r_1, r_2, \ldots, r_n]$ denotes the corresponding full reasoning trajectory consisting of $n$ intermediate steps (\textit{i.e.}, thoughts), and $\mathbf{y}$ is the final answer. 
Following \cite{zhang2025lightthinker}, we use ``{\textbackslash}n{\textbackslash}n'' as a delimiter to simply segment the CoT trajectory $\mathbf r$ into different thoughts $\{ r_i \}$. 
In addition, let $\boldsymbol{\theta}$ denote the parameters of an LLM, and let $\pi_{\boldsymbol{\theta}}(\mathbf{b} \mid \mathbf{a})$ represent the model’s predicted probability of generating response $\mathbf{b}$ given input $\mathbf{a}$.
The reasoning training for $\btheta$ under $\mathcal D$ can be then cast as 
\begin{align}
\begin{array}{ll}
 \displaystyle \minimize_{\boldsymbol{\theta}} & -\mathbb{E}_{(\mathbf{x}, \mathbf{r}, \mathbf{y}) \in \mathcal{D}} \left[ \log \pi_{\boldsymbol{\theta}}(\mathbf{r}, \mathbf{y} \mid \mathbf{x}) \right],
\end{array}
\label{eq:sft}
\end{align}
where the training objective is defined as a cross-entropy-based sequence prediction loss, which maximizes the likelihood of generating the reasoning trace and final answer conditioned on the input.

\paragraph{Prior work: Efficiency-accuracy trade-off through dataset size reduction.}
While SFT on long CoT significantly enhances the reasoning abilities of LLMs, it is highly resource-intensive, particularly when the traces are generated by LRMs like \textsc{DeepSeek-R1}. This renders solving problem \eqref{eq:sft} computationally expensive, particularly in resource-constrained settings such as academic labs.

\begin{wrapfigure}{r}{0.35\textwidth}
 \vspace*{-1mm}
\centering
\includegraphics[width=0.35\textwidth]{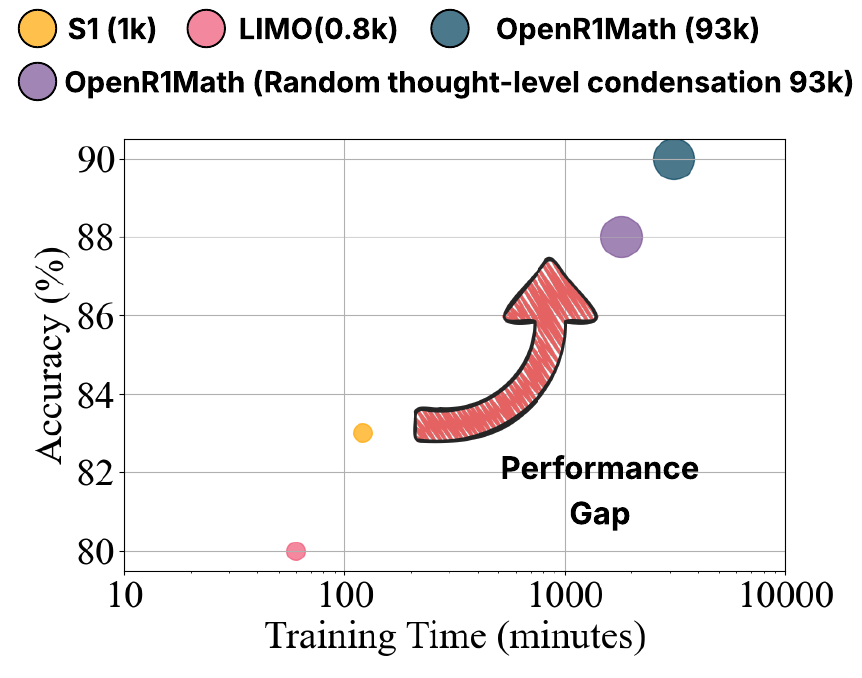} 
\vspace*{-8mm}
  \caption{\footnotesize{Accuracy and training time for reasoning training on OpenR1Math (93k examples), LIMO (0.8k examples), and S1 (1k examples), using \textsc{Qwen2.5-Math-7B-Instruct} as the base non-reasoning LLM. Accuracy is evaluated on the MATH500 benchmark. In addition to standard CoT datasets, we also include a thought-level condensed version of OpenR1Math, where 50\% of the intermediate thoughts in each CoT trace are randomly retained and the remainder pruned.
  }}
  \vspace*{-6mm}
  \label{fig:different_dataset}
\end{wrapfigure}

To improve the efficiency of reasoning training, prior work has explored size-reduced, high-quality CoT datasets such as S1 \citep{muennighoff2025s1} and LIMO \citep{ye2025limo}, each containing around 1k carefully curated examples. However, we find that these datasets are typically benchmarked on large models (\textit{e.g.}, 32B), and their effectiveness does not consistently transfer to the training of smaller models.
As shown in \textbf{Figure\,\ref{fig:different_dataset}}, training a 7B model on LIMO or S1 significantly speeds up reasoning training compared to conventional SFT using the larger CoT dataset OpenR1Math (93k examples). However, 
this speedup comes at the cost of reduced accuracy on MATH500, 80.0\% and 83.6\% for LIMO and S1, respectively, compared to 90.2\% achieved when training on OpenR1Math.
This indicates that small-scale datasets like S1 and LIMO are insufficient to consistently support effective reasoning performance.

\paragraph{Problem statement.}
As motivated by Figure\,\ref{fig:different_dataset}, curating smaller CoT datasets does not appear to be an effective solution for improving the efficiency of reasoning training while preserving reasoning performance.
Therefore, we propose shifting the focus from \textit{reducing the number of training examples} to \textit{condensing the reasoning trajectory within each example}. That is, we ask whether \textit{thought-level} condensation of CoT data, rather than example-level dataset reduction, can enable more efficient and effective reasoning training.
Therefore, we define the CoT condensation operation as the selection (or pruning) of intermediate thoughts within a reasoning trajectory. Given a CoT trace $\mathbf{r} = [r_1, r_2, \ldots, r_n]$, the condensed version is denoted as $\mathbf{r}_{\mathrm{cond}} = [r_{i}]_{i \in \Omega}$, where $\Omega \subseteq \{1, \dots, n\}$ is the index set of selected thoughts and the remaining thoughts are discarded.
The potential of thought-level CoT condensation is evident with random thought selection. As shown in \textbf{Figure\,\ref{fig:different_dataset}}, randomly selecting 50\% of the reasoning steps in each example from OpenR1Math yields 88.0\% accuracy, outperforming LIMO and S1, while cutting training time by approximately 40\% compared to training on the full OpenR1Math.

Figure\,\ref{fig:different_dataset} motivates the central research question of our work: \textit{Can we design an effective CoT condensation method to address the supervised fine-tuning problem in \eqref{eq:sft}, one that substantially reduces training cost while preserving reasoning performance comparable to full-length CoT supervision?}

%% file: sections/method_SL.tex
\section{Edge-Preserving CoT Condensation: Method and Rationale}
\label{sec: method}
\begin{wrapfigure}{r}{0.49\textwidth}
 \vspace*{-4mm}
\centering
\includegraphics[width=0.49\textwidth]{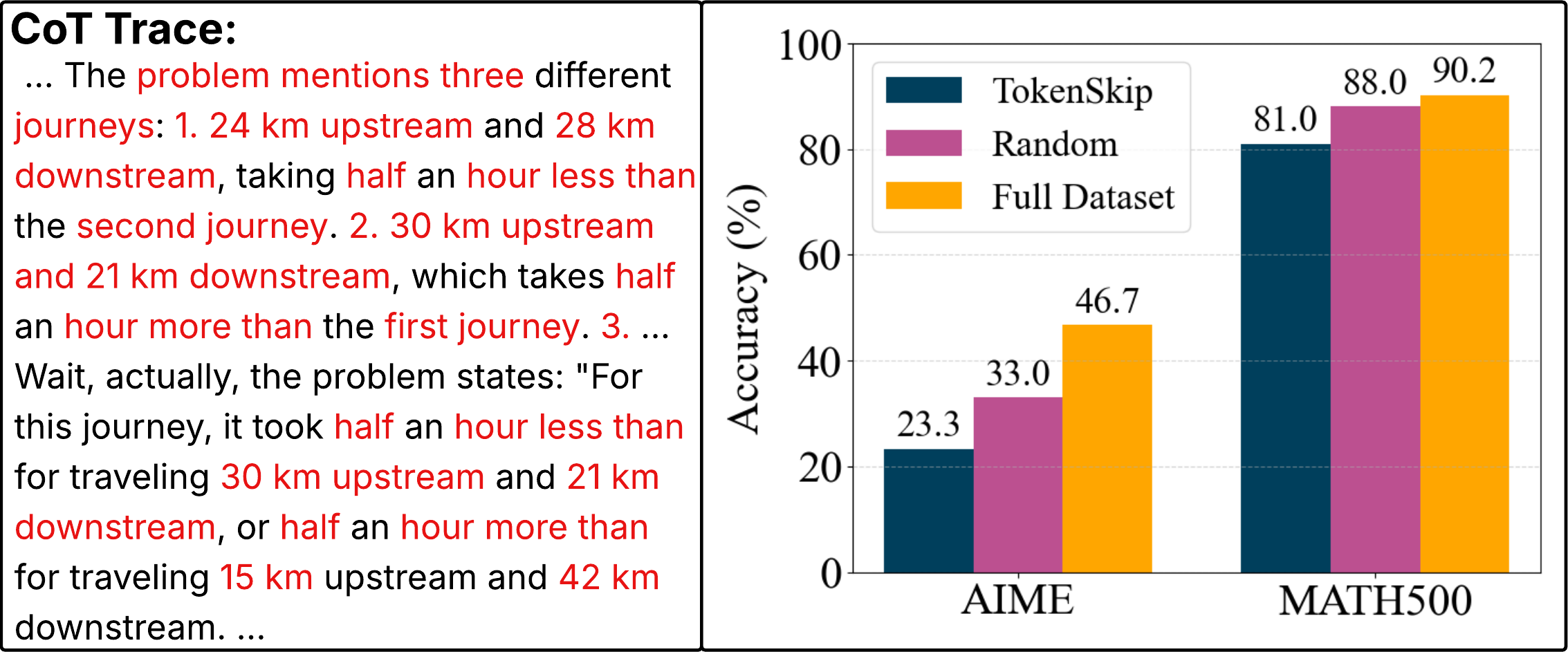}
\vspace*{-5mm}
  \caption{\footnotesize{{
  {Performance of TokenSkip-based token-level condensation for reasoning training. \textit{(Left)} Visualization of a CoT trace pruned by TokenSkip \cite{xia2025tokenskip} with a 50\% pruning ratio. Tokens highlighted in \textcolor{red}{red} are retained, while the rest are removed. \textit{(Right)} Final answer accuracy of models trained on three datasets: TokenSkip-pruned (50\%), random thought-level condensation (50\%), and the original full dataset, evaluated on AIME and MATH500. All models are fine-tuned from \textsc{Qwen2.5-Math-7B-Instruct} on the OpenR1Math dataset.}
  } }
  } 
  \vspace*{-7mm}
  \label{fig:token_prune}
\end{wrapfigure}
In this section, we begin with a warm-up study to motivate why individual thoughts (\textit{i.e.}, reasoning steps) serve as a proper unit for condensing CoT traces. We then present our method, \ours{}, which performs CoT condensation by leveraging the underlying structure of reasoning trajectories. Finally, we justify the design of \ours{} from two complementary perspectives: (1) the mutual information between individual reasoning steps and the final answer, and (2) a sensitivity analysis that contrasts the importance of reasoning structure versus content.


\paragraph{CoT condensation unit: Thoughts or tokens?}  
As an alternative to thought-level condensation, one may consider pruning a CoT trace at the token level. A representative approach is \textbf{TokenSkip} \cite{xia2025tokenskip} to assign importance scores to individual tokens and prune those deemed less critical \cite{pan2024llmlingua}. We can apply this method to compress CoT traces. However, when training models on these token-pruned CoT datasets, we observe a {significant} drop in performance compared to training on the original, unpruned data, as shown in \textbf{Figure\,\ref{fig:token_prune}}.
This may be because token-level pruning disrupts the thought-level reasoning pattern. For example, as shown in  Figure\,\ref{fig:token_prune}(Left), TokenSkip overlooks transitional markers and reflective words (\textit{e.g.}, ``wait'') that are essential for connecting thoughts and preserving logical flow.
Therefore, token-level condensation becomes {\textit{in}effective}  for reasoning training, 
performing even worse than random thought-level condensation in  Figure\,\ref{fig:token_prune}(Right).

\begin{wrapfigure}{r}{0.33\textwidth}
 \vspace*{-4mm}
\centering
\includegraphics[width=0.33\textwidth]{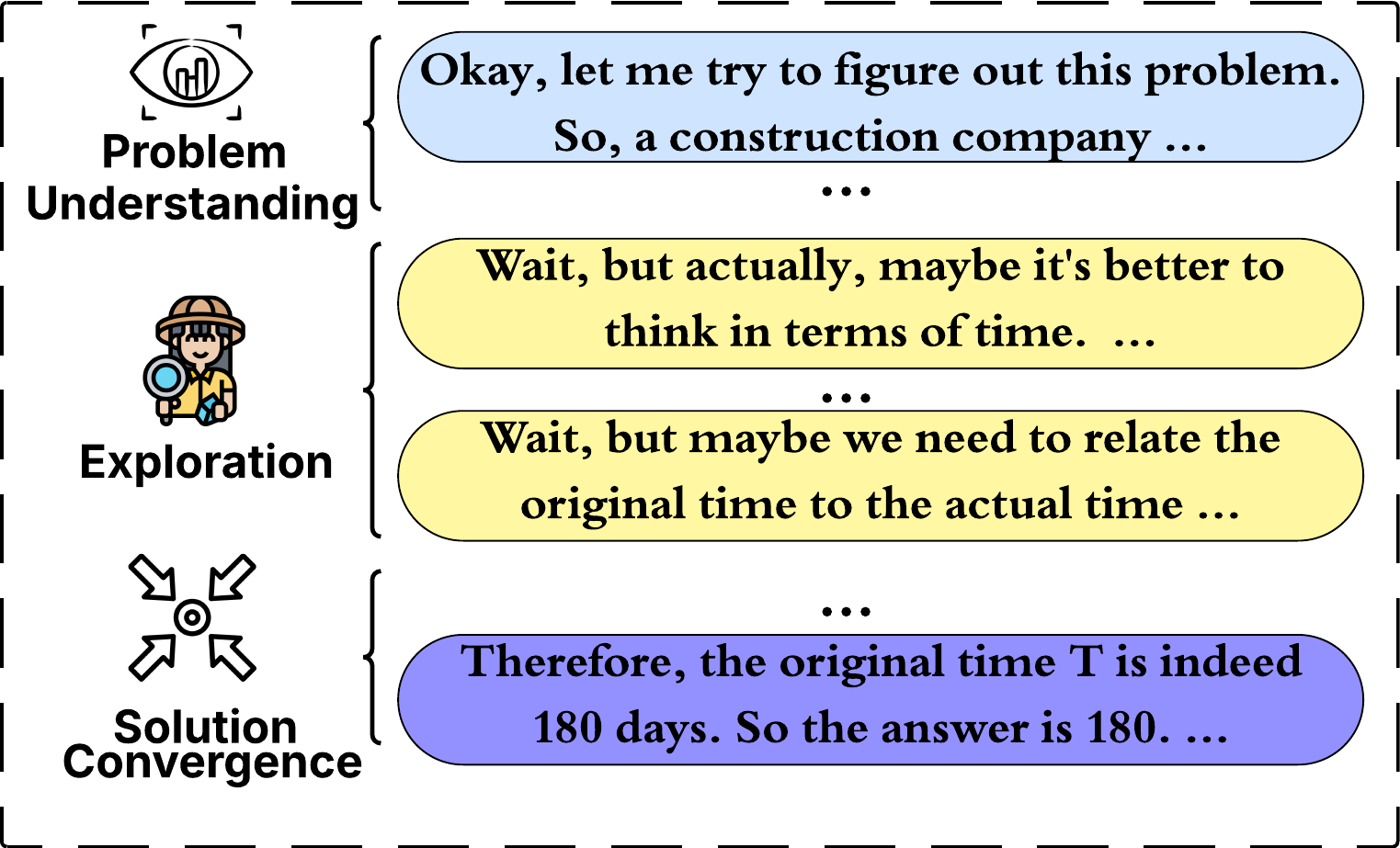} 
\vspace*{-5mm}
  \caption{\footnotesize{Illustration of three-stage structure of long CoT reasoning: problem understanding (head), exploration (middle), and solution convergence (tail).}}
  \vspace*{-5mm}
  \label{fig:think_stages}
\end{wrapfigure}
\paragraph{Edge-preserving condensation ({\ours}).}
First, we segment each CoT trace into three distinct stages based on their functions: \textit{(1) Understand}: parsing and interpreting the problem; \textit{(2) Explore}: inferring and iterating through reasoning paths; And \textit{(3) Converge}: synthesizing information and finalizing the solution. See \textbf{Figure\,\ref{fig:think_stages}} for an illustration. These stages correspond to the head, middle, and tail segments of the CoT trace, respectively. Based on this structured 
progression, we can achieve reasoning condensation by selectively removing one of these stages from the full CoT trace.

Based on the above CoT segmentation, 
we next develop \ours{}, a method that preserves only the head and tail portions of the CoT trace, effectively connecting the initial and final stages while discarding the exploration stage. This design mirrors the idea of retaining the ``edges'' of a reasoning trajectory.  Recall that $\mathbf{r} = [r_1, r_2, \dots, r_n]$ denotes the full reasoning trajectory consisting of $n$ thoughts. We define the \textit{condensation ratio} (\textbf{CR}) $\tau \in [0, 1]$ as the fraction of thoughts retained after pruning (\textit{i.e.}, the length of the condensed trajectory).  
\ours{} compresses the full reasoning trajectory $\mathbf{r}$ into a condensed version $\mathbf{r}_{\mathrm{cond}}$ by pruning the middle portion of the CoT trace while retaining the proper head and tail segments:
\begin{align}
\mathbf{r}_{\mathrm{cond}} &= [r_i]_{i \in \Omega}, \quad 
\Omega = \left\{1, \dots, \left\lfloor \tfrac{\tau n}{2} \right\rfloor \right\} \cup \left\{n - \left\lfloor \tfrac{\tau n}{2} \right\rfloor + 1, \dots, n\right\}.
\tag{\ours{}}
\label{eq: Epic}
\end{align}
Here, $\lfloor \cdot \rfloor$ denotes the floor function. The total number of retained thoughts, $\lfloor \tau n \rfloor$, is equally divided between the head and tail segments, each of length $\lfloor \tfrac{\tau n}{2} \rfloor$. Please refer to Appendix~\ref{app: cond_data_ex} for visualizations of example reasoning traces after condensation.

\begin{wrapfigure}{r}{0.45\textwidth}
 \vspace*{-7mm}
\centering
\includegraphics[width=0.45\textwidth]{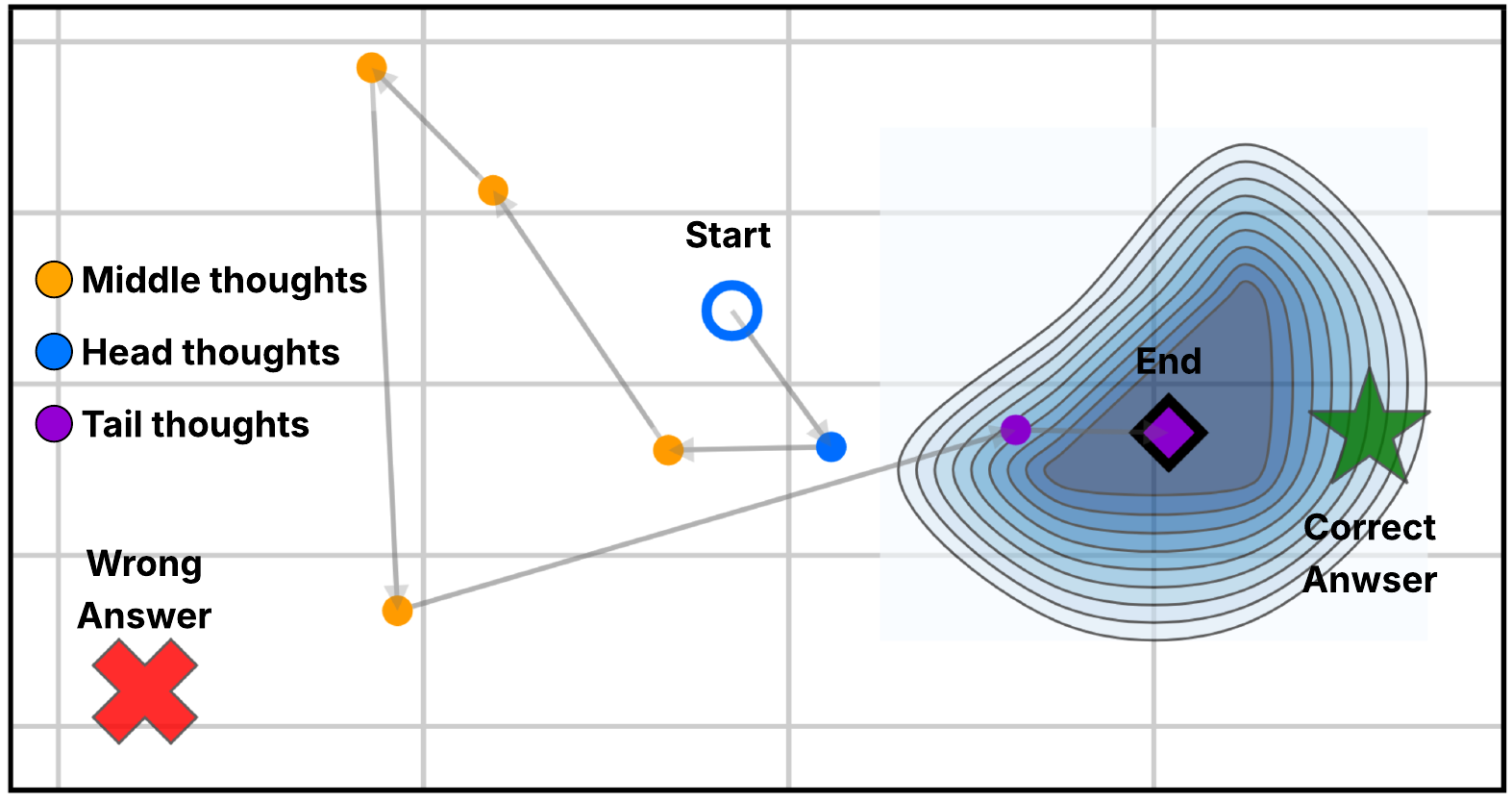} 
\vspace*{-5mm}
  \caption{\footnotesize{
Visualization of a reasoning trajectory generated by \textsc{DeepSeek-R1-Distill-Qwen-7B} on the AQuA \citep{ling2017program} dataset. The plot is produced using the trajectory landscape tool from \citep{zhou2025landscape}, where each node represents the model’s reasoning state in a latent space after $k$ thought steps. And $x$ and $y$ axes correspond to two t-SNE-projected dimensions. The trajectory is segmented into three parts: the first 25\% of steps (\textcolor{blue}{blue}, ``head thoughts"), the middle 50\% (\textcolor{orange}{orange},``middle thoughts"), and the final 25\% (\textcolor{violet}{violet},``tail thoughts"). The correct answer is shown as a green star, while red cross denotes incorrect (distractor) answers. 
  }}
  \vspace*{-7mm}
  \label{fig:loss_landscape}
\end{wrapfigure}

Although \ref{eq: Epic} appears simple, its effectiveness can be demonstrated through extensive empirical results in {Sec.\,\ref{sec: exp}}: It enables efficient reasoning training on condensed CoT datasets while preserving performance comparable to full-length supervision.
In what follows, we validate the soundness of \ours{} using a CoT landscape visualization tool \cite{zhou2025landscape}, offering an interpretable view of how individual reasoning steps contribute to final answer generation.
This stems from our observation that the exploration stage (\textit{i.e.}, the middle portion discarded by \ours{}), which is often the longest part of the reasoning trace, can be redundant or even distracting sometimes.

Leveraging the visualization tool from \cite{zhou2025landscape}, \textbf{Figure\,\ref{fig:loss_landscape}} illustrates the landscape of a reasoning trajectory projected into a latent semantic space, where darker regions represent intermediate reasoning states that are semantically \textit{closer} to the correct answer. The $x$ and $y$ axes correspond to two t-SNE-projected dimensions. The visualization shows that the initial and final reasoning steps tend to contribute more directly to the final answer, as reflected by their proximity to the correct solution. In contrast, a large portion of intermediate steps often drift away from this path, potentially steering the model toward suboptimal conclusions. This supports our hypothesis that the middle portion of a CoT trace may be less informative, or even detrimental, to accurate reasoning.

\paragraph{Understanding {\ours} via mutual information (MI).}
To further understand which parts of the reasoning trajectory are most important for improving reasoning ability, we analyze {\ours} using MI. Our goal is to quantify how much information different portions of the reasoning trace retain compared with the full reasoning trace. For a condensed trace $\mathbf{r}_{\mathrm{cond}} = [r_i]_{i \in \Omega}$, we obtain a matrix representation $\mathbf{E}_{\Omega} = [\mathbf{e}^{\Omega}_1, \dots, \mathbf{e}^{\Omega}_m]^\top \in \mathbb{R}^{m \times d}$ by feeding each trace through a pretrained LLM and applying mean pooling over the final hidden states across the token dimension. Here, $m$ denotes the number of samples used for MI evaluation and $d$ is the hidden dimension of the model. 
{We compute the mutual information between $\mathbf{E}_{\Omega}$ and $\mathbf{E}_{\mathrm{Full}}$, denoted as $\mathcal{I}(\mathbf{E}_{\Omega}; \mathbf{E}_\mathrm{Full})$, using the Kraskov estimator \cite{kraskov2004estimating}, which approximates MI based on distances between nearest neighbors in the sample space. See {Appendix\,\ref{app: mi}} for more details. The MI score serves as a proxy for how informative the selected reasoning steps are compared with the full reasoning trace.}



\begin{wraptable}{r}{0.4\textwidth}
\vspace*{-2mm}
\caption{\footnotesize{Comparison of MI, computed using \eqref{eq: MI}, between the full reasoning trajectory and selected portions of the reasoning trajectory under various condensation methods and condensation ratios ($\tau$). The evaluation is performed on 2500 examples sampled from the OpenR1Math dataset using the \textsc{Qwen2.5-1.5B-Instruct} model.}}
\resizebox{\linewidth}{!}{ 
\begin{tabular}{c|cccc}
\toprule
\midrule
\textbf{Method}
& 
$\tau=0.01$&$\tau=0.05$ & $\tau=0.1$ & $\tau=0.5$
\\
\midrule
Full ($\tau = 1$)   & \multicolumn{4}{c}{8.77} \\
\midrule
Random & 0.56 & 1.90 & 2.64 & 4.57 \\
\HoC    & 0.93 & 1.77 & 2.27 & 4.85 \\
\MoC    & 0.64 & 1.39 & 1.84 & 3.81 \\
\ToC    & 0.43 & 1.06 & 1.46 & 3.05 \\
\rowcolor{blue!20}
\ours   & 3.07 & 3.57 & 4.06 & 8.70  \\
\midrule
\bottomrule
\end{tabular}
}
\vspace{-6mm}
\label{tab:MI}
\end{wraptable}
A higher $\mathcal{I}(\mathbf{E}_{\Omega}; \mathbf{E}_\mathrm{Full})$ indicates that the condensed subset $\Omega$ preserves more of the information in the full reasoning trace, and thus corresponds to a more effective condensation strategy. We compute MI between the full reasoning trace and different portions of the reasoning trajectory to assess the informativeness of each segment. Specifically, given a CR (condensation ratio) $\tau$,  we define: \textit{Head-only Condensation} (\textbf{\HoC}) as $\Omega_{\mathrm{H}} = \left\{1, \dots, \left\lfloor \tau n \right\rfloor \right\}$, \textit{Tail-only Condensation} (\textbf{\ToC}) as $\Omega_{\mathrm{T}} = \left\{n - \left\lfloor \tau n \right\rfloor + 1, \dots, n \right\}$, and \textit{Middle-only Condensation} (\textbf{\MoC}) as $\Omega_{\mathrm{M}} = \left\{ \left\lfloor \tfrac{(1 - \tau) n}{2} \right\rfloor + 1, \dots, n - \left\lfloor \tfrac{(1 - \tau) n}{2} \right\rfloor \right\}$. As shown in \textbf{Table\,\ref{tab:MI}}, {\ours} consistently achieves the highest MI across all condensation ratios $\tau \in \{0.01, 0.05, 0.1, 0.5\}$, closely matching the MI of the full reasoning trace. This indicates that {\ours} effectively preserves the most informative parts of the reasoning trace. Notably, at $\tau = 0.5$, EPiC attains an MI of 8.70, nearly matching the full trace MI of 8.77. This indicates that EPiC preserves nearly all the semantic content of the full reasoning trajectory while using only half the tokens, providing strong evidence that its structural selection strategy captures the most informative parts of the trace. We observe similar results using \textsc{Qwen2.5-Math-7B-Instruct}, as shown in Table\,\ref{tab:MI_7b}.

\begin{wrapfigure}{r}{0.39\textwidth}
\vspace*{-5mm}
\centering
\includegraphics[width=0.39\textwidth]{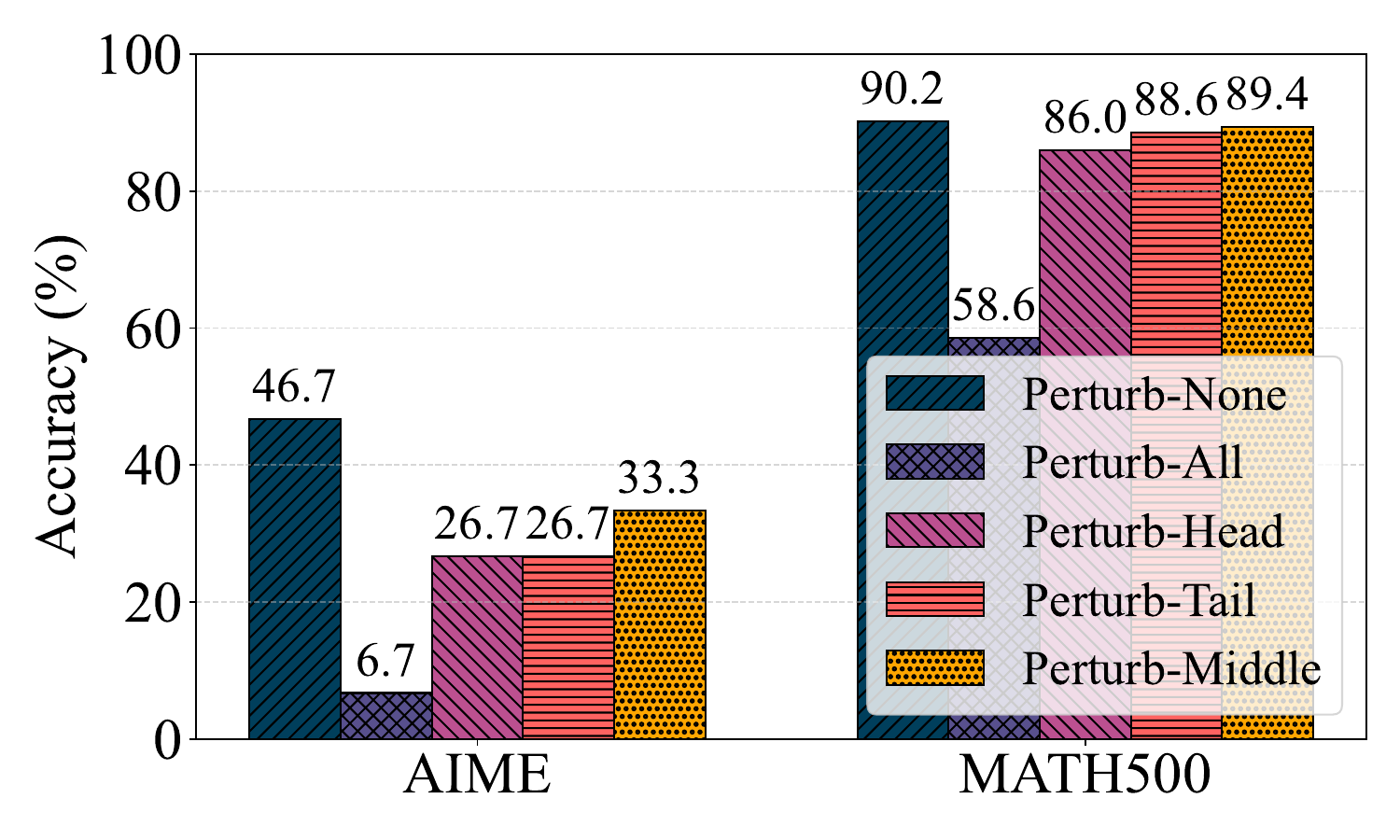} 
\vspace*{-7mm}
  \caption{\footnotesize{
  Final answer accuracy comparison for reasoning training using \textsc{Qwen2.5-Math-7B-Instruct} on various perturbed CoT training sets, evaluated on AIME and MATH500 at test time. Perturbations are applied to specific regions of the CoT trace--head, tail, middle, entire trace-or not applied at all (no perturbation).
  }}
  \vspace*{-6mm}
  \label{fig:random_perturb}
\end{wrapfigure}
\paragraph{Reasoning structure may matter more than content.} 
%
To assess the importance of different CoT stages, we conduct a perturbation analysis inspired by \ours{}. Rather than removing the reasoning steps not in the selected index set $\Omega$, we replace their content with randomly sampled text while preserving the structural layout of the full trace. When perturbing the reasoning content, we preserve reflection tokens, often realized as discourse markers such as \texttt{wait} and \texttt{hmm} \citep{muennighoff2025s1}, as they serve as important transitional and reflective cues that help maintain coherence between thoughts.


Building on the above, we investigate the impact of fixing the reasoning condensation pattern $\Omega$ while perturbing the unselected thoughts $\{ r_i \}_{i \notin \Omega}$ by replacing their content (between reflection tokens) with randomly sampled sentences from WikiText \cite{merity2016pointer}. 
\textbf{Figure\,\ref{fig:random_perturb}} shows how reasoning training on the perturbed CoT dataset impacts model performance. We evaluate five settings: perturbing (1) all reasoning steps, or 50\% of the trace located in the (2) head, (3) middle, or (4) tail segments. Perturbing the middle region results in the smallest performance degradation, achieving 89.4\% accuracy on MATH500, compared to 90.2\% with the original dataset. In contrast, perturbing the head or tail leads to more significant accuracy drops, while perturbing the entire trace severely degrades performance. These results support the core idea of \ours{}: The middle stage of a CoT trace is less critical than the head and tail, and much of its content can be pruned or abstracted without substantially compromising reasoning ability.

%% file: sections/experiments.tex
\section{Experiments}
\label{sec: exp}
\subsection{Experiment Setups}
\paragraph{Training datasets.}
To demonstrate the effectiveness of {\ours} in facilitating CoT training for enhanced reasoning capabilities, we train models on two long-form CoT datasets distilled from \textsc{DeepSeek-R1}: \textbf{(1) OpenR1Math} \cite{openr1}: This dataset comprises 220k math problems, each paired with reasoning traces generated by \textsc{DeepSeek-R1}. Answers are verified using either a math verifier \cite{math-verify} or \textsc{LLaMA-3.3-70B-Instruct} to ensure correctness. In our experiments, we use the default main subset, which includes 93k verified examples. \textbf{(2) GeneralThoughts} \cite{generalthought195k}: This dataset offers a diverse reasoning traces beyond mathematics and coding, spanning natural sciences, humanities, social sciences, and general conversational reasoning. The traces are generated by a diverse set of strong LLMs, including \textsc{O3-Mini}, \textsc{Gemini-2-Flash-Thinking}, \textsc{Claude-3.7-Sonnet}, and \textsc{DeepSeek-R1}.

\paragraph{Model setups.}
In our experiments, we primarily use {the \textit{non-reasoning} LLM} \textsc{Qwen2.5-Math-7B-Instruct} \cite{yang2024qwen2-math} as the base model for SFT-based reasoning training, due to its strong mathematical capabilities. To evaluate the robustness and generalizability of {\ours} across different model initializations, we additionally conduct experiments with two alternative models: \textsc{Qwen2.5-7B-Instruct} \cite{yang2024qwen2}, which shares the same architecture but lacks math-specific instruction tuning, and \textsc{LLaMA3.1-8B-Instruct} \cite{grattafiori2024llama}, which differs in both architecture and pretraining corpus. These variants assess {\ours}'s effectiveness when initialized from a weaker math model or a different architecture.

\paragraph{Evaluation benchmarks.}
To assess the acquired reasoning capabilities, we primarily evaluate models on three benchmarks: \textbf{(1) MATH500} \cite{lightman2023let}: A curated set of 500 multi-step problems from the OpenAI MATH benchmark, designed to measure mathematical reasoning ability. \textbf{(2) AIME24} \cite{aime}: A set of 30 high school competition-level math problems from the 2024 American Invitational Mathematics Examination (AIME). \textbf{(3) GPQA-Diamond} \cite{rein2024gpqa}: A graduate-level STEM benchmark consisting of multiple-choice questions in biology, physics, and chemistry. All problems are written and verified by domain experts (PhD-level), providing a challenging testbed for evaluating general scientific reasoning beyond mathematics.
For evaluation, we set a maximum generation length of 9000 tokens for both \textsc{MATH500} and \textsc{AIME24}, and 4000 tokens for \textsc{GPQA-Diamond}. Decoding is performed using nucleus sampling with a temperature of 0.6 and top-$p$ of 0.95, following \cite{guo2025deepseek}. {In addition to final answer accuracy, we also assess reasoning generation quality using two auxiliary metrics: (1) the length of the generated reasoning traces, and (2) the number of reflection tokens, which serve as strong indicators of reasoning ability, \textit{e.g.}, the ``Aha Moment''  emphasized in \textsc{DeepSeek-R1} \cite{guo2025deepseek}. }


\paragraph{Baselines.}
To evaluate the effectiveness of {\ours}, we compare it against several baseline condensation strategies: \textbf{(1) Random:} Randomly selects a subset of reasoning steps based on the same condensation ratio $\tau$, without considering their position within the trajectory. \textbf{(2) {\HoC} (Head-only Condensation):} Retains only the first $\lfloor \tau n \rfloor$ steps of the reasoning trajectory, where $n$ is the total number of steps. \textbf{(3) {\ToC} (Tail-only Condensation):} Retains only the last $\lfloor \tau n \rfloor$ steps. \textbf{(4) TokenSkip} \cite{xia2025tokenskip}: A recent token-level condensation that scores and selects important tokens across the trace for retention. {Unless otherwise specified, the condensation ratio $\tau$ is set to 50\% throughout our experiments.} Please refer to the training and implementation details in Appendix\,\ref{app: train_details}.

\subsection{Experiment Results}

\begin{table}[htb]
\centering
\caption{\footnotesize{
Performance comparison of {\ours} against full dataset training and baseline condensation methods across three reasoning benchmarks: {Math500}, {AIME24}, and {GPQA-Diamond}. Each benchmark reports both accuracy (\%) , the average number of generated tokens (\#Toks) and the average number of reflection tokens (\#Rtoks).
All models are trained via SFT from {Qwen2.5-Math-7B-Instruct}, using a fixed condensation ratio of 50\%. The reasoning training is conducted on two datasets, \textsc{OpenR1Math} and \textsc{GeneralThought195k}, respectively. The final column reports the total training time in hours.
}}
\label{tab:multi_dataset_comparison}
\resizebox{0.85\textwidth}{!}{\begin{tabular}{l |ccc|ccc|ccc|c}
\toprule
\textbf{Methods} & 
\multicolumn{3}{c|}{\textbf{Math500}} & 
\multicolumn{3}{c|}{\textbf{AIME24}} & 
\multicolumn{3}{c|}{\begin{tabular}{c}
   \textbf{GPQA}-\textbf{Diamond}
\end{tabular}} & 
\textbf{Time} \\
 & Acc & \#Toks & \#RToks & Acc & \#Toks& \#RToks & Acc & \#Toks & \#RToks & (Hours) \\
\midrule
w/o SFT & 82.6 & 696.6 & 0.0 & 3.3 & 1624.2 & 0.1 & 34.9 & 1331.5 & 0.0 & - \\
\midrule
\multicolumn{11}{c}{\textbf{OpenR1Math} \citep{openr1}} \\
\midrule
Full dataset & 90.2 & 3213.5  & 17.6 & 46.7 & 7365.2& 43.7 & 38.4 & 3817.1 & 37.0 & 51.9 \\
\midrule
Tokenskip    & 81.0 & 4861.5 & 28.0 & 23.3 & 8499.4 & 48.3 & 31.3 & 3896.7&30.4 & 30.0 \\
Random       & 88.0 & 3221.9& 17.2& 33.3 & 7382.8& 45.8  & 36.4 & 3802.7 &37.8& 32.4 \\
\HoC         & 89.6 & 3178.3 & 17.3 & 33.3 & 7549.0 & 45.9& 40.4 & 3753.5 & 39.5&34.2 \\
\ToC         & 84.6 & 3088.6 &16.4 & 33.3 & 7141.4 & 41.8 & 43.9 & 3472.8 & 28.7&32.0 \\
\rowcolor{blue!20}
\ours        & 90.2 & 3109.1&17.5 & 40.0 & 7330.8&45.4 & 41.9 & 3725.7  & 37.8&34.0 \\
\midrule
\multicolumn{11}{c}{\textbf{GeneralThought195k} \citep{generalthought195k}} \\
\midrule
Full dataset & 87.0 & 3072.7 & 23.3 & 26.7 & 7613.6&50.2 & 40.4 & 3494.9&46.2  & 48.8 \\
\midrule
Tokenskip    & 58.4 & 4281.8 &0.0& 0.0 & 9000.0&0.0 & 29.3 & 3653.5&0.0  & 32.0 \\
Random       & 58.2 & 3621.8 &27.0& 0.0& 7591.9&52.0 & 35.0 & 3329.4&41.4  & 32.0 \\
\HoC         & 85.8 & 3252.7 &18.3& 26.7 & 8004.9&43.8 & 37.4 & 3490.1 &40.1& 33.5 \\
\ToC         & 75.4 & 2963.8&19.1 & 13.3 & 6991.3 &46.6& 41.4 & 3182.5 &31.5& 31.6 \\
\rowcolor{blue!20}
\ours        & 86.0 & 2874.2 &18.5& 20.0 & 7967.2&46.9 & 42.4 & 3388.3 &40.8& 32.3 \\
\bottomrule
\end{tabular}}
\vspace*{-5mm}
\end{table}

\paragraph{Performance overview of {\ours} vs. full-data training and condensation baselines.} 
In \textbf{Table,\ref{tab:multi_dataset_comparison}}, we evaluate the performance of {\ours} under a 50\% condensation ratio across three reasoning benchmarks: \textsc{Math500}, \textsc{AIME24}, and \textsc{GPQA-Diamond}. We compare against baselines trained on two datasets: \textsc{OpenR1Math}, which focuses exclusively on mathematical problems, and \textsc{GeneralThought195k}, which contains reasoning traces spanning diverse domains such as science, humanities, and general knowledge. All models are fine-tuned from \textsc{Qwen2.5-Math-7B-Instruct}.

{First}, {\ours} matches the performance of full-data training while significantly reducing training time up to \textit{34\%}. For example, when trained on \textsc{OpenR1Math}, {\ours} achieves 90.2\% accuracy on \textsc{Math500}, identical to the full model, but requires only {34.0 hours} of training compared to {51.9 hours} for the full dataset. On \textsc{GeneralThought195k}, {\ours} also maintains comparable performance (86.0\% vs. 87.0\%) while reducing training time from {48.8 to 32.3 hours}, demonstrating substantial efficiency gains without performance loss.


In-domain evaluation on \textsc{GPQA-Diamond} further validates the strength of {\ours}. Since \textsc{GeneralThought195k} includes STEM-related reasoning (\textit{e.g.}, physics and biology), \textsc{GPQA-Diamond} serves as a natural in-domain test. As shown, {\ours} achieves {42.4\%} accuracy, outperforming the full-data baseline of 40.4\%. Even when treated as an out-of-domain task--training only on math-focused \textsc{OpenR1Math}--{\ours} generalizes better than full-data training (41.9\% vs. 38.4\%), suggesting that pruning the middle portion of reasoning traces may help improve generalization.


Compared to other structural condensation baselines, {\ours} also exhibits clear advantages. Both {\HoC} and {\ToC}, which preserve only the head or tail of the reasoning trace, perform noticeably worse than {\ours} across all tasks and datasets. These results confirm that preserving both the beginning and end of the reasoning trace, while discarding the middle, is a highly effective and efficient strategy.

Last but not least, we analyze reasoning behavior through the lens of generation length. Across all benchmarks, {\ours} produces responses of \textit{comparable length} to those generated by models trained on the full dataset, indicating that it effectively preserves reasoning complexity. For example, on \textsc{Math500}, {\ours} achieves the same 90.2\% accuracy as the full model while generating, on average, only 100 fewer tokens (3109.1 vs. 3213.5). In addition, we observe that {\ours} maintains a similar number of reflection tokens, such as "wait", "hmm", and other metacognitive markers, compared to the full model, further confirming that its condensed traces still elicit rich and deliberate reasoning behavior during inference.
This result is particularly noteworthy given that a 50\% condensation ratio was applied to the CoT training data. Despite this reduction, the model’s ability to generate complete and coherent reasoning traces at test time remains essentially \textit{lossless} compared to full-data training.
Appendix\,\ref{app: examples_eval} also includes qualitative generation examples illustrate that {\ours} achieves comparable reasoning quality to the full-data model.

\begin{figure*}[htb]
\vspace*{-0mm}
\centerline{
\begin{tabular}{cccc}
    \hspace*{-2mm}  \includegraphics[width=0.23\textwidth,height=!]{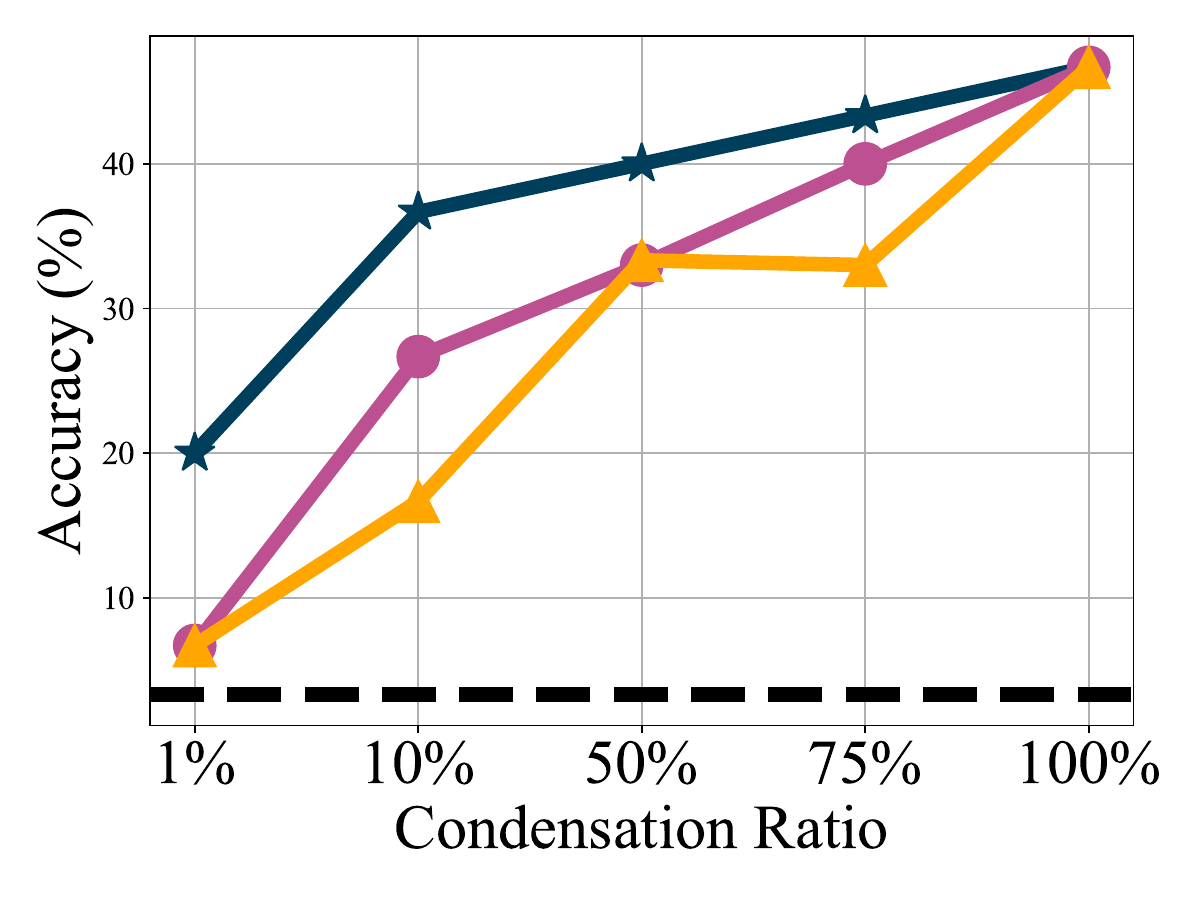} &
    \hspace*{-5mm} \includegraphics[width=0.23\textwidth,height=!]{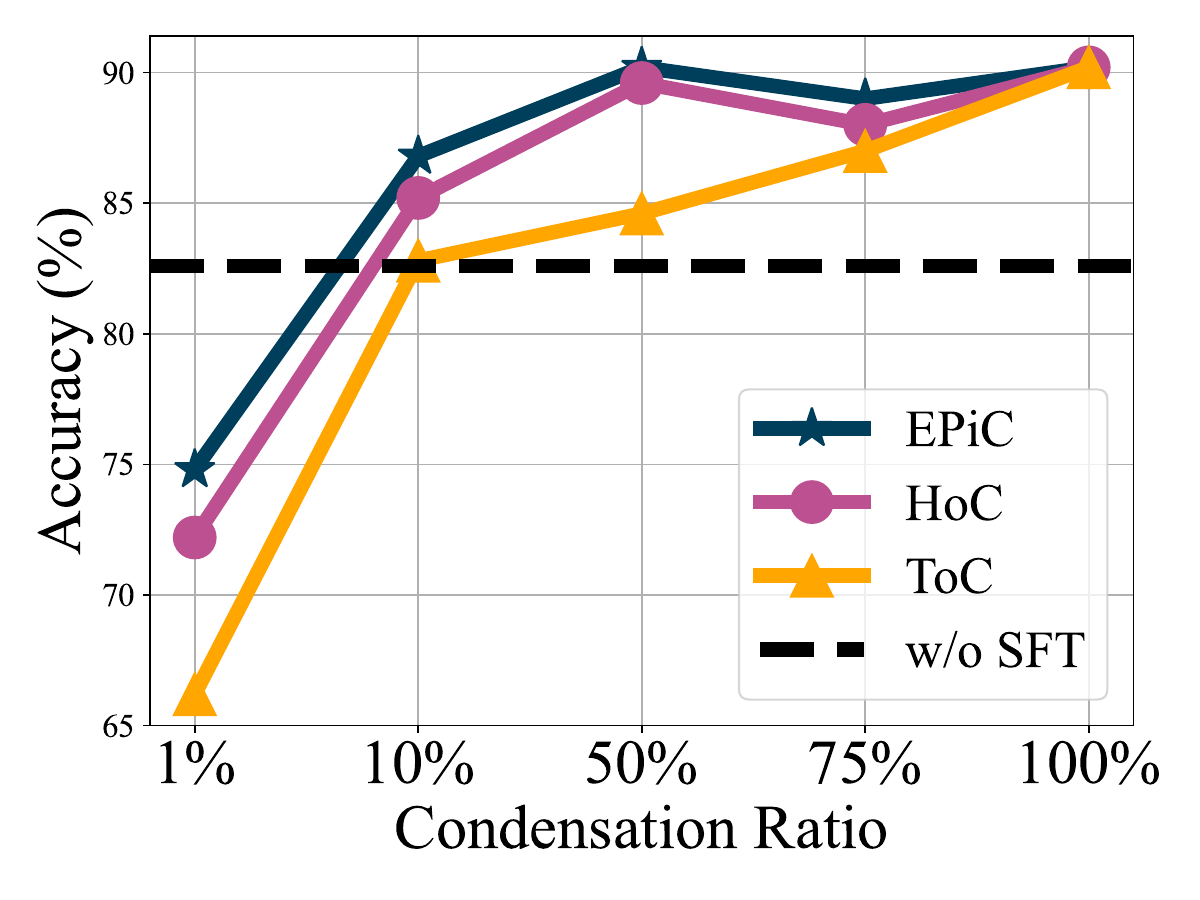} &
    \hspace*{-5mm} \includegraphics[width=0.23\textwidth,height=!]{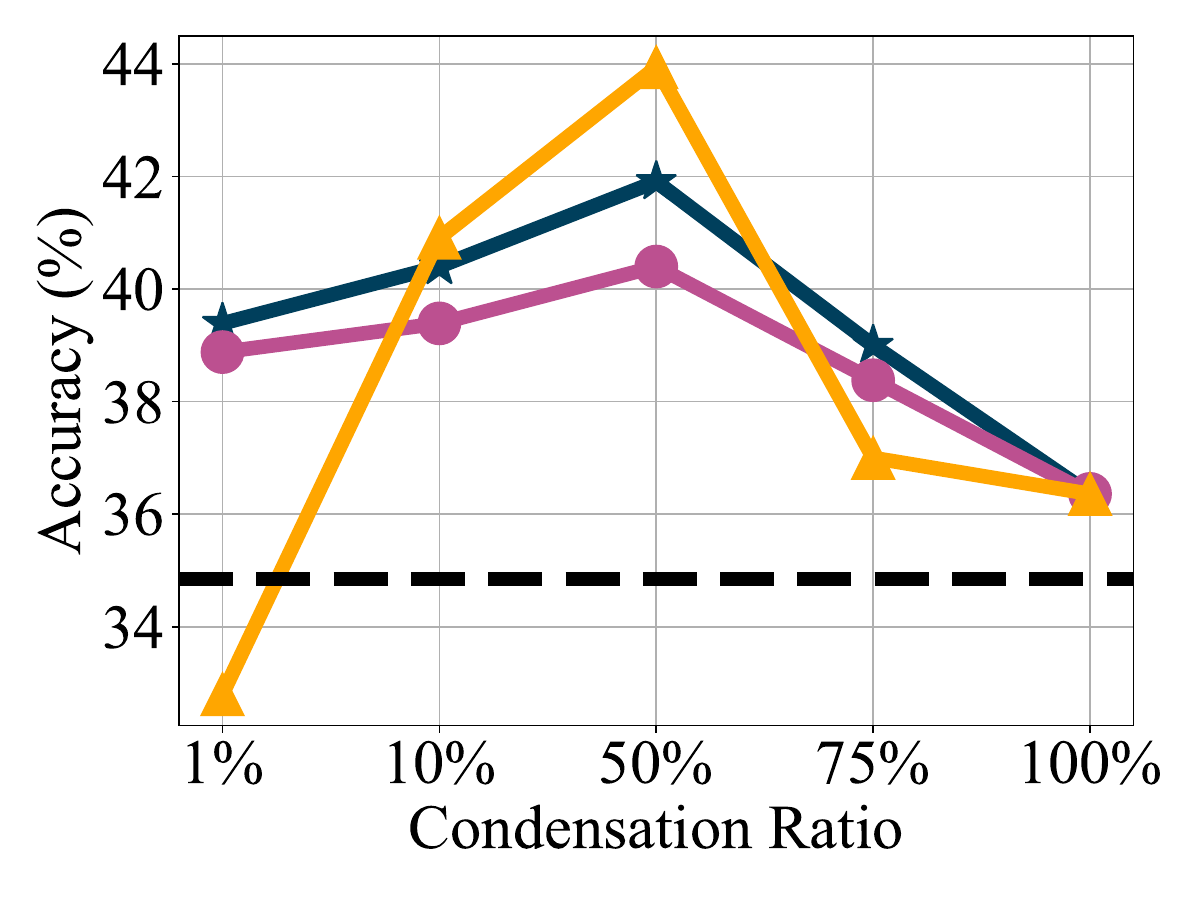}     &\includegraphics[width=0.28\textwidth,height=!]{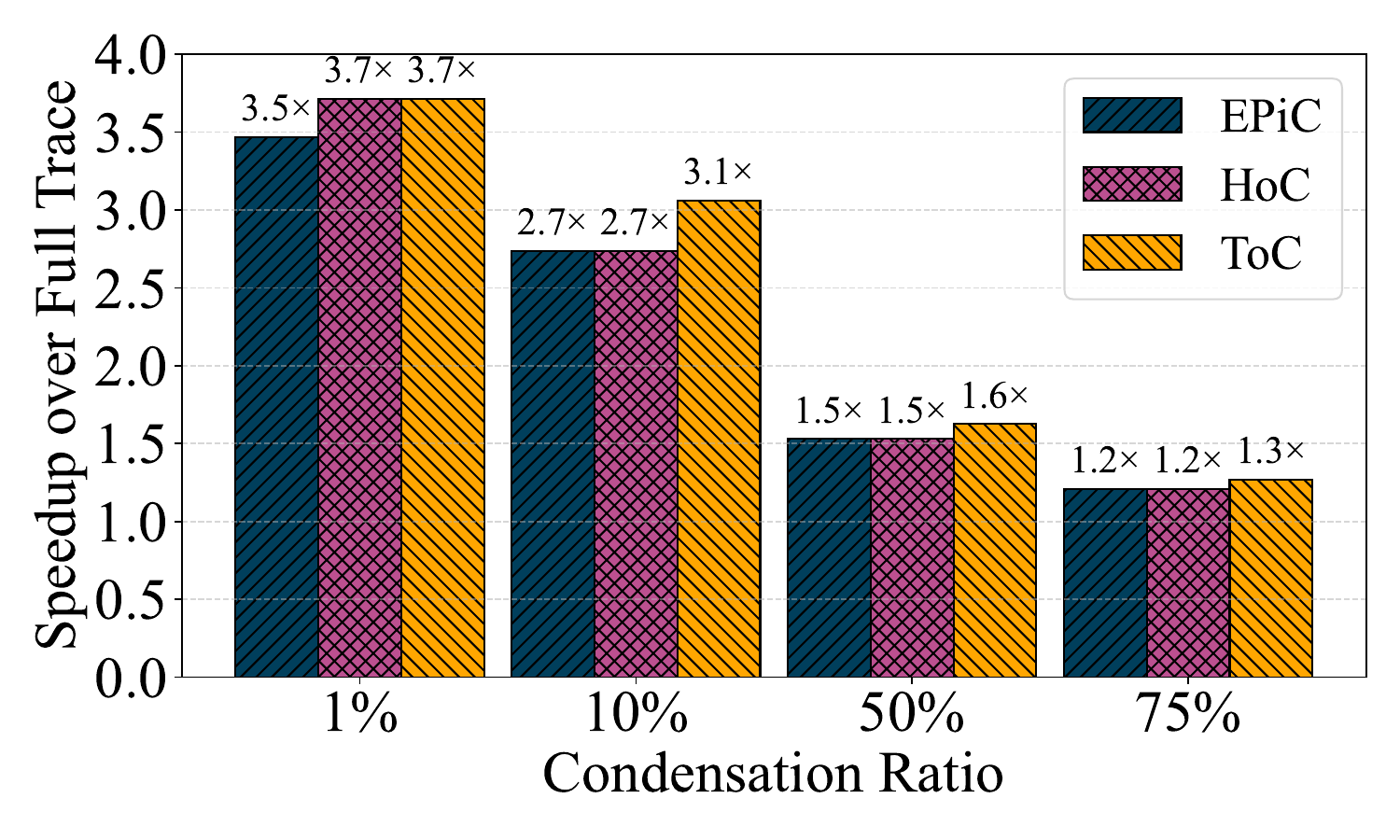}   \vspace*{-1mm} \\
    \footnotesize{(a) AIME24} & \footnotesize{(b) MATH500} & (c) \footnotesize{GPQA-Diamond} & (d) \footnotesize{Speedup} 
\end{tabular}}
 \vspace*{-2mm}
\caption{\footnotesize{Reasoning accuracy of CoT training at different condensation ratios using {\ours}, {\HoC}, and {\ToC}, on three benchmarks: 
\textbf{(a)} AIME24, 
\textbf{(b)} MATH500, and 
\textbf{(c)} GPQA-Diamond. 
All models are fine-tuned using {Qwen2.5-Math-7B-Instruct} on OpenR1Math. 
The dashed line indicates performance without SFT, and the 100\% condensation ratio refers to the full training dataset. \textbf{(d)} shows the training speedup relative to full-trace fine-tuning across condensation ratios for each method.
}}
\vspace*{-3mm}
\label{fig: ratio_reason}
\end{figure*}



%
\paragraph{Performance against CoT condensation ratios.} In \textbf{Figure\,\ref{fig: ratio_reason}-(a,b,c)}, we present the performance of reasoning training under varying CoT condensation ratios using different condensation methods ({\ours}, {\HoC}, {\ToC}), evaluated on the reasoning benchmarks {AIME24}, {MATH500}, and {GPQA-Diamond}.
As expected, reasoning performance generally improves as the condensation ratio increases (\textit{i.e.}, more thoughts are retained), where 100\% corresponds to the full-data training scenario. Notably, {GPQA-Diamond} exhibits a sweet spot at 50\% condensation, where the accuracy even surpasses that of the full-dataset baseline. This suggests that moderate pruning may help eliminate noisy or redundant reasoning, thereby improving generalization.
We also observe that the use of CoT training data plays a key role in driving reasoning accuracy, compared to the initial model without SFT.
Across all benchmarks and condensation levels, {\ours} consistently outperforms both {\HoC} and {\ToC}, reinforcing the results shown in Table\,\ref{tab:multi_dataset_comparison}. Furthermore, \textbf{Figure\,\ref{fig: ratio_reason}-(d)} reports the corresponding training speedups relative to full CoT fine-tuning. As expected, lower condensation ratios lead to faster training. At 50\% condensation, {\ours} delivers a substantial {1.5$\times$} improvement in training efficiency over full-data training.

\begin{figure}[h]
\centering
\includegraphics[width=0.39\textwidth]{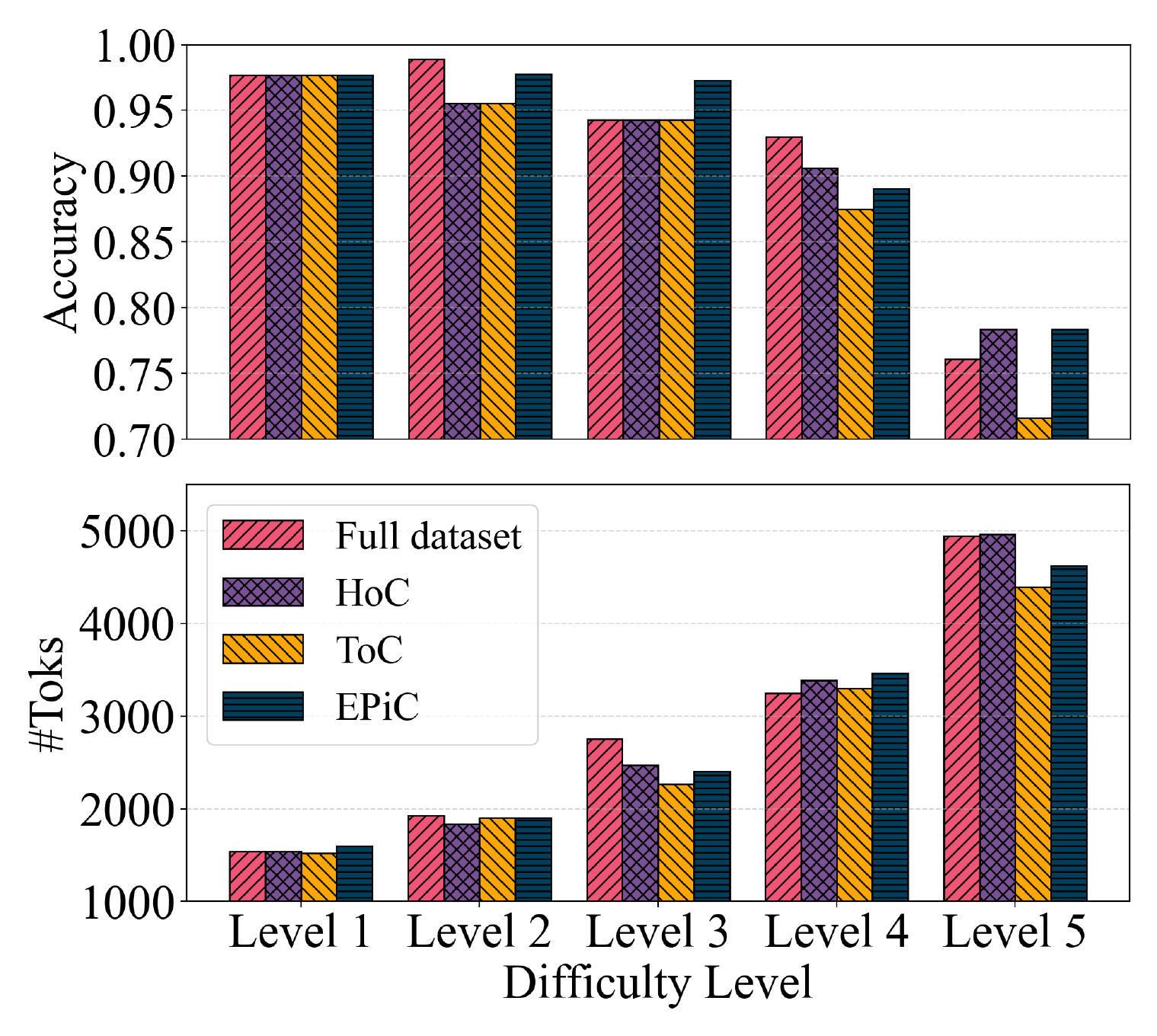} 
\vspace*{-2mm}
\caption{\footnotesize{
Accuracy and generation length across difficulty levels on the {Math500} benchmark.
\textbf{(Top)} Accuracy comparison of condensation methods ({50\%} condensation ratio) and full-data baseline across five difficulty levels.
\textbf{(Bottom)} Average number of generated tokens per method and difficulty level.
All models are fine-tuned from \textsc{Qwen2.5-Math-7B-Instruct} on  {OpenR1Math}.
}}
  \vspace*{-5mm}
  \label{fig:diff_acc_tok}
\end{figure}
\paragraph{Performance against problem difficulty levels.} In \textbf{Figure\,\ref{fig:diff_acc_tok}}, we present a fine-grained breakdown of model performance across five difficulty levels on the {Math500} benchmark. The top plot shows accuracy comparisons, while the bottom plot reports the corresponding average number of generated tokens. All models are trained using \textsc{Qwen2.5-Math-7B-Instruct} on the {OpenR1Math} dataset.
As expected, accuracy generally decreases and generation length increases as problem difficulty rises. This trend holds consistently across all condensation strategies, reflecting the intrinsic complexity of harder problems. However, 
 {\ours} performs \textit{better than the full-data baseline} on the most difficult {Level 5} problems. This indicates that CoT condensation via {\ours}
did \textit{not} disproportionately disadvantage on the harder levels. 
In addition, as evidenced by the bottom plots, all condensation methods applied to the CoT training datasets do not hinder the reasoning generation capability of the resulting models after training, across all problem difficulty levels. This also echoes the finding in Table\,\ref{tab:multi_dataset_comparison} that reasoning ability can be effectively acquired using shorter CoT traces without compromising generation quality.

\paragraph{{\ours} is resilient to non-reasoning base model choice.}
To evaluate the robustness of {\ours} under different initialization conditions, we assess its performance when fine-tuning two distinct pretrained backbones: \textsc{Qwen2.5-7B-Instruct} \cite{yang2024qwen2} and \textsc{LLaMA3.1-8B-Instruct} \cite{grattafiori2024llama}. As shown in \textbf{Table\,\ref{tab:diff_models}}, {\ours} consistently achieves strong performance despite using only 50\% of the original reasoning traces. Compared to full-data fine-tuning, it attains comparable or even superior accuracy while significantly reducing training time, saving {19.1 hours} on LLaMA3 and {19 hours} on Qwen. These results indicate that {\ours} generalizes well  across base models.
\begin{table}[h]
\centering
\caption{\footnotesize{Performance comparison between full-dataset training and {\ours} on the OpenR1Math dataset across different backbone models, with similar format as Table~\ref{tab:multi_dataset_comparison}.
}}
\label{tab:diff_models}
\resizebox{.59\textwidth}{!}{\begin{tabular}{l |cc|cc|cc|c}
\toprule
\textbf{Methods} & 
\multicolumn{2}{c|}{\textbf{Math500}} & 
\multicolumn{2}{c|}{\textbf{AIME24}} & 
\multicolumn{2}{c|}{\begin{tabular}{c}
   \textbf{GPQA}  \\
   \textbf{Diamond}
\end{tabular}} & 
\textbf{Time} \\
 & Acc & \#Toks & Acc & \#Toks & Acc & \#Toks & (Hours) \\
\midrule
\multicolumn{8}{c}{\textsc{Qwen2.5-7B-Instruct}} \\
\midrule
w/o SFT & 76.4 & 583.0  & 10.0 & 1061.8  & 30.8 & 577.8 & - \\
Full dataset & 84.4  &  3499.4 & 26.7 & 7590.1 & 35.9 & 3798.7 & 52.1  \\
\rowcolor{blue!20}
\ours        & 84.2 & 3378.4 & 26.7 & 7839.1 & 35.9 & 3760.1  & 33.2  \\
\midrule
\multicolumn{8}{c}{\textsc{LLaMA3.1-8B-Instruct}} \\
\midrule
w/o SFT &  47.4 & 1311.7 & 3.3 & 3180.6 & 29.8 & 1060.4 & - \\
Full dataset & 78.2 & 9000.0  & 13.3 & 9000.0 & 27.2 &  4000.0 & 56.6 \\
\rowcolor{blue!20}
\ours        & 75.0 & 9000.0 & 16.7 & 9000.0 & 29.8 & 4000.0  & 37.5 \\
\bottomrule
\end{tabular}}
\vspace*{-5mm}
\end{table}

%% file: sections/conclusion.tex
\vspace*{-1mm}
\section{Conclusion}
\vspace*{-1mm}
In this work, we investigate a novel perspective on reasoning supervision by proposing \textbf{\ours}, a simple yet effective thought-level condensation strategy that preserves only the \textit{head} and \textit{tail} portions of long CoT traces. Motivated by the observation that intermediate reasoning steps often contain redundant or exploratory content, we systematically evaluate the informativeness of different reasoning segments using mutual information analysis. Our findings reveal that removing the middle portion of CoT traces leads to significant efficiency gains without sacrificing model performance. Extensive experiments across multiple reasoning benchmarks demonstrate that {\ours} achieves comparable or even superior performance to full-dataset training, while reducing training time. The method proves robust across datasets, difficulty levels, and model architectures, offering a practical and interpretable solution for improving the efficiency of reasoning fine-tuning. Efficient reasoning supervision is increasingly important as the cost of full CoT training becomes prohibitive. Limitations and broader impacts are further discussed in \textbf{Appendix\,\ref{appendix: limitation}} and \textbf{Appendix\,\ref{appendix: impact}}.
\section*{Acknowledgments}
The work of J. Jia, C. Fan, and
S. Liu is supported by the National Science Foundation (NSF) CISE Core Program Award IIS-2207052, the
NSF CAREER Award IIS-2338068, the Cisco Faculty Research Award,   the ARO Award W911NF2310343, and the Amazon Research Award for AI in Information Security.  

%% file: sections/appendix.tex
\appendix
\setcounter{section}{0}

\section*{Appendix}

\setcounter{section}{0}
\setcounter{figure}{0}
\makeatletter 
\renewcommand{\thefigure}{A\arabic{figure}}
\renewcommand{\theHfigure}{A\arabic{figure}}
\renewcommand{\thetable}{A\arabic{table}}
\renewcommand{\theHtable}{A\arabic{table}}

\makeatother
\setcounter{table}{0}

\setcounter{mylemma}{0}
\renewcommand{\themylemma}{A\arabic{mylemma}}
\setcounter{equation}{0}
\renewcommand{\theequation}{A\arabic{equation}}
 
\section{Visualization of Condensed Reasoning Examples}
To provide qualitative insight into how {\ours} condense long reasoning traces, Table~\ref{tab:exp_train_data} presents an example from the condensed training dataset based on \textsc{OpenR1Math}. This example highlights the head and tail portions retained by {\ours}, with the pruned middle segment shown in red. 
\label{app: cond_data_ex}
\begin{table}[htb]
    \centering
\caption{\footnotesize{Visualization of a training example with \ours{}. $\mathbf{x}$ is the input question, $\mathbf{r}$ is the full CoT reasoning trace, and $\mathbf y$ is the final answer. The \textcolor{red}{red}-highlighted reasoning steps in $\mathbf{r}$ correspond to the middle portion that is pruned, while the head and tail segments are retained for training.}}
   \resizebox{0.75\textwidth}{!}{ \begin{tabular}{>{\centering\arraybackslash}m{0.5cm}|>{\raggedright\arraybackslash}m{12.0cm}}
    $\mathbf{x}$ & \texttt{A ship traveling along a river has covered \$24 km\$ upstream and \$28 km\$ downstream, ... Determine the speed of the ship in still water and the speed of the river.} \\
    \midrule
    $\bfr$ & \textbf{<think>}\texttt{\textbackslash nOkay, so I need to find the speed of the ship in still water and the speed of the river. Let me start by recalling that when a ship is moving upstream, its effective speed is the speed of the ship minus the speed of the river. Conversely, when moving downstream, its effective speed is the ship's speed plus the river's speed. \textbackslash n\textbackslash nLet me denote the speed of the ship in still water as \textbackslash ( v \textbackslash ) (in km/h) and the speed of the river as \textbackslash ( r \textbackslash ) (also in km/h). Then, the upstream speed would be \textbackslash ( v - r \textbackslash ), and the downstream speed would be \textbackslash ( v + r \textbackslash ).\textbackslash n\textbackslash nThe problem mentions three different journeys:\textbackslash n\textbackslash n1. 24 km upstream and 28 km downstream, taking half an hour less than the second journey.\textbackslash n2. ...} \texttt{\textcolor{red}{Let me first work on the first equation:\textbackslash n\textbackslash n\textbackslash ( \textbackslash frac\{24\}\{v - r\} + \textbackslash frac\{28\}\{v + r\} = \textbackslash frac\{30\}\{v - r\} + \textbackslash frac\{21\}\{v + r\} - 0.5 \textbackslash )\textbackslash n\textbackslash nLet me subtract \textbackslash ( \textbackslash frac\{24\}\{v - r\} + \textbackslash frac\{28\}\{v + r\} \textbackslash ) from both sides, but maybe it's better to bring all terms to the left side:\textbackslash n\textbackslash n\textbackslash ( \textbackslash frac\{24\}\{v - r\} + \textbackslash frac\{28\}\{v + r\} - \textbackslash frac\{30\}\{v - r\} - \textbackslash frac\{21\}\{v + r\} + 0.5 = 0 \textbackslash )\textbackslash n\textbackslash nSimplify the terms:\textbackslash n\textbackslash nFor the upstream terms: \textbackslash ( 24 - 30 = -6 \textbackslash ), so \textbackslash ( \textbackslash frac\{-6\}\{v - r\} \textbackslash )\textbackslash n\textbackslash nFor the downstream terms: \textbackslash ( 28 - 21 = 7 \textbackslash ), so \textbackslash ( \textbackslash frac\{7\}\{v + r\} \textbackslash )\textbackslash n\textbackslash nSo, the equation becomes:\textbackslash n\textbackslash n\textbackslash ( \textbackslash frac\{-6\}\{v - r\} + \textbackslash frac\{7\}\{v + r\} + 0.5 = 0 \textbackslash )\textbackslash n\textbackslash nSimilarly, for the second equation:\textbackslash n\textbackslash n\textbackslash ( \textbackslash frac\{24\}\{v - r\} + \textbackslash frac\{28\}\{v + r\} - \textbackslash frac\{15\}\{v - r\} - \textbackslash frac\{42\}\{v + r\} - 0.5 = 0 \textbackslash )\textbackslash n\textbackslash nSimplify:\textbackslash n\textbackslash nUpstream terms: \textbackslash ( 24 - 15 = 9 \textbackslash ), so \textbackslash ( \textbackslash frac\{9\}\{v - r\} \textbackslash )\textbackslash n\textbackslash nDownstream terms: \textbackslash ( 28 - 42 = -14 \textbackslash ), so \textbackslash ( \textbackslash frac\{-14\}\{v + r\} \textbackslash )\textbackslash n\textbackslash nThus:\textbackslash n\textbackslash n\textbackslash ( \textbackslash frac\{9\}\{v - r\} - \textbackslash frac\{14\}\{v + r\} - 0.5 = 0 \textbackslash )\textbackslash n\textbackslash nNow, we have two equations:\textbackslash n\textbackslash n1. \textbackslash ( -\textbackslash frac\{6\}\{v - r\} + \textbackslash frac\{7\}\{v + r\} + 0.5 = 0 \textbackslash )\textbackslash n2. \textbackslash ( \textbackslash frac\{9\}\{v - r\} - \textbackslash frac\{14\}\{v + r\} - 0.5 = 0 \textbackslash ) ...}} \texttt{... \textbackslash ( 2v = 20 \textbackslash ) \u21d2 \textbackslash ( v = 10 \textbackslash )\textbackslash n\textbackslash nSubstituting back into \textbackslash ( v + r = 14 \textbackslash ):\textbackslash n\textbackslash n\textbackslash ( 10 + r = 14 \textbackslash ) \u21d2 \textbackslash ( r = 4 \textbackslash )\textbackslash n\textbackslash nSo, the speed of the ship in still water is 10 km/h, and the speed of the river is 4 km/h.\textbackslash n\textbackslash nLet me verify these results by plugging them back into the original time equations.\textbackslash n\textbackslash nFirst journey: 24/(10-4) + 28/(10+4) = 24/6 + 28/14 = 4 + 2 = 6 hours.\textbackslash n\textbackslash nSecond journey: 30/(10-4) + 21/(10+4) = 30/6 + 21/14 = 5 + 1.5 = 6.5 hours. So, 6.5 - 0.5 = 6 hours, which matches the first journey. Good.\textbackslash n\textbackslash nThird journey: 15/(10-4) + 42/(10+4) = 15/6 + 42/14 = 2.5 + 3 = 5.5 hours. Then, 5.5 + 0.5 = 6 hours, which also matches the first journey. Perfect.\textbackslash n\textbackslash nTherefore, the solution is correct.\textbackslash n\textbackslash n**Final Answer**\textbackslash nThe speed of the ship in still water is \textbackslash boxed\{10\} km/h and the speed of the river is \textbackslash boxed\{4\} km/h.\textbackslash n} \textbf{</think>}\\
    \midrule
        $\mathbf{y}$ & \textbf{***Fianl Answer***}\texttt{, ..., The speed of the ship in still water is \textbackslash(\textbackslash \text{boxed}\{10\}\textbackslash) km/h and the speed of the river is \textbackslash(\textbackslash \text{boxed}\{4\}\textbackslash) km/h.}
    \end{tabular}
}
    \label{tab:exp_train_data}
\end{table}

\section{Implementation Details of Mutual Information and Extended  Analysis}
\label{app: mi}
\begin{wraptable}{r}{0.40\linewidth}
\vspace*{-7mm}
\caption{\footnotesize{Comparison of MI, computed using \eqref{eq: MI}, between the full reasoning trajectory and selected portions of the reasoning trajectory under various condensation methods and condensation ratios ($\tau$). The evaluation is performed on 2500 examples sampled from the OpenR1Math dataset using the \textsc{Qwen2.5-7B-Instruct} model.}}
\resizebox{\linewidth}{!}{ 
\begin{tabular}{c|cccc}
\toprule
\midrule
\textbf{Method}
& 
$\tau=0.01$&$\tau=0.05$ & $\tau=0.1$ & $\tau=0.5$
\\
\midrule
Full ($\tau = 1$)   & \multicolumn{4}{c}{8.79} \\
\midrule
Random & 0.41 & 1.79 & 2.48 & 4.27 \\
\HoC    & 1.03 & 1.97 & 2.39 & 4.90 \\
\MoC    & 0.41 & 1.25 & 1.71 & 3.65 \\
\ToC    & 0.46 & 1.19 & 1.50 & 3.07 \\
\rowcolor{blue!20}
\ours   & 3.11 & 3.58 & 4.07 & 8.67  \\
\midrule
\bottomrule
\end{tabular}
}
\vspace{-5.5mm}
\label{tab:MI_7b}
\end{wraptable}
\paragraph{Implementation details of mutual information.} To identify which parts of the reasoning trajectory contribute most to model learning, we analyze the mutual information (MI) between different portions of the trace and the full CoT trajectory. For a condensed trace $\mathbf{r}_{\mathrm{cond}} = [r_i]_{i \in \Omega}$, we encode it using a pretrained LLM and apply mean pooling over the token dimension to obtain a representation matrix $\mathbf{E}_{\Omega} \in \mathbb{R}^{m \times d}$, where $m$ is the number of samples and $d$ is the hidden dimension. We then compute MI between $\mathbf{E}_{\Omega}$ and the full trace embedding $\mathbf{E}_{\mathrm{Full}}$ as $\mathcal{I}(\mathbf{E}_{\Omega}; \mathbf{E}_\mathrm{Full})$ using the Kraskov $k$-nearest neighbor estimator~\cite{kraskov2004estimating}, which approximates MI based on distances between nearest neighbors in the sample space. This non-parametric method is well-suited for high-dimensional representations and provides a robust estimate of MI without requiring density assumptions. We use $k = 5$ in all experiments. For each $i \in \{1, \ldots, m\}$, we compute the radius $\rho_i$ using the $\ell_\infty$ norm as
\begin{align*}
    \rho_i = \min_{j \in \mathcal{E}_{i,k}} \max \left\{ \| \mathbf{e}_i^{\Omega} - \mathbf{e}_j^{\Omega} \|_{\infty}, \| \mathbf{e}_i^{\mathrm{Full}} - \mathbf{e}_j^{\mathrm{Full}} \|_{\infty} \right\},
\end{align*}
where $\mathbf{e}_i^{\Omega}$ and $\mathbf{e}_i^{\mathrm{Full}}$ are the $i$th rows of $\mathbf{E}_{\Omega}$ and $\mathbf{E}_{\mathrm{Full}}$, respectively.
$\mathcal{E}_{i,k} \subseteq \{1, \ldots, m\} \setminus \{i\}$ denotes the indices of the $k$-nearest neighbors of the joint embedding $(\mathbf{e}_i^{\mathrm{Full}}, \mathbf{e}_i^{\Omega})$ in the joint space $\mathbb{R}^{2d}$. Using this radius $\rho_i$, we then count the number of neighbors of $\mathbf{e}_i^{\Omega}$ and $\mathbf{e}_i^{\mathrm{Full}}$ that lie within $\rho_i$ in their respective marginal spaces
\begin{align*}
    n^{\Omega}_i = \left|\left\{ j \neq i\ ;\ \| \mathbf{e}_i^{\Omega} - \mathbf{e}_j^{\Omega} \|_{\infty} < \rho_i \right\}\right|, \quad
    n^{\mathrm{Full}}_i = \left|\left\{ j \neq i\ ;\ \| \mathbf{e}_i^{\mathrm{Full}} - \mathbf{e}_j^{\mathrm{Full}} \|_{\infty} < \rho_i \right\}\right|.
\end{align*}
Finally, we estimate the mutual information as
\begin{align}
\mathcal{I}(\mathbf{E}_{\Omega}; \mathbf{E}_\mathrm{Full}) = \psi(k) + \psi(m) - \frac{1}{m} \sum_{i=1}^{m} \left[ \psi(n^{\Omega}_i + 1) + \psi(n^{\mathrm{Full}}_i + 1) \right],
\label{eq: MI}
\end{align}
where $\psi(\cdot)$ is the digamma function. The MI score \eqref{eq: MI} serves as a proxy for how informative the selected reasoning steps are compared with the full reasoning trace.

\paragraph{Additional results for mutual information.} To further validate the robustness of our mutual information analysis, we perform an additional evaluation using the model, \textsc{Qwen2.5-7B-Instruct}, to compute the latent representations $\mathbf{E}_{\Omega} $. As shown in Table\,\ref{tab:MI_7b}, {\ours} consistently achieves the highest MI across all tested condensation ratios $\tau \in \{0.01, 0.05, 0.1, 0.5\}$, outperforming all other condensation baselines. Remarkably, at $\tau = 0.5$, {\ours} attains an MI of 8.67, which is almost indistinguishable from the MI of the full reasoning trace (8.79). These findings are consistent with the results reported in Table\,\ref{tab:MI}, and further corroborate that {\ours} preserves the majority of semantic content in the reasoning trace while using only 50\% of the tokens. This highlights the effectiveness of our method in maintaining reasoning fidelity under significant token budget constraints.

\section{Additional Experimental Details}
\label{app: train_details}
\subsection{Training setup}
\noindent \textbf{Supervised fine-tuning setup.}  
We adopt a unified training configuration across all base models and data condensation strategies to ensure fair comparison. All models are fine-tuned for 3 epochs using the AdamW optimizer with a learning rate of $5 \times 10^{-5}$, weight decay of 0.0001, and a linear learning rate scheduler with 10\% warmup. Training is performed on 8 NVIDIA A6000 GPUs with a global batch size of 16, achieved via a per-device batch size of 1 and gradient accumulation over 2 steps. We use \texttt{bfloat16} precision and enable gradient checkpointing for memory efficiency. To improve throughput, long sequences are packed into fixed-length inputs with a maximum context length of 32,768 tokens.
\subsection{Inference setup}
For evaluation, we set a maximum generation length of 9000 tokens for both \textsc{MATH500} and \textsc{AIME24}, and 4000 tokens for \textsc{GPQA-Diamond}. Decoding is performed using nucleus sampling with a temperature of 0.6 and top-$p$ of 0.95, following~\cite{guo2025deepseek}. For \textsc{AIME24}, we sample 32 responses per query and report \textit{pass@1}. For all other benchmarks, we report accuracy from a single sampled response.

\section{Visualizations of Model Responses after Training with {\ours}}
\label{app: examples_eval}
To qualitatively evaluate the effectiveness of {\ours}, we present representative examples of model-generated responses from the \textsc{AIME24} benchmark in Table\,\ref{tab:aime_example}. The example is generated by a model fine-tuned using the condensed dataset produced by {\ours}. In the visualization, the input question (\textbf{x}), the reasoning trace (\textbf{r}), and the final answer (\textbf{y}) are shown. We highlight reflection cues and structural tokens (\textit{e.g.}, \texttt{Wait}, \texttt{Therefore}, \texttt{<think>}, \texttt{</think>}) in \textcolor{red}{\textbf{red}} to indicate the model's internal reasoning structure. The correct final answer is marked in \textcolor{green}{green}. The visualization demonstrates that despite being trained on pruned traces that remove the middle portion of reasoning steps, the model retains coherent reasoning abilities and produces accurate answers, without hurting its capacity for reflection or disrupting its thinking structure. This supports that {\ours} enables reasoning enhancement without compromising the model’s reflective capacity or overall reasoning ability.
\begin{table}[htb]
    \centering
    \caption{\footnotesize{Visualization of a model-generated response from the AIME24 benchmark, produced by a model fine-tuned using {\ours}. \textbf{x} denotes the input question, \textbf{r} the generated reasoning trace, and \textbf{y} the final answer. \textcolor{red}{\textbf{Red}} highlights reflection cues (\textit{e.g.}, {Wait}, {Therefore}) and structural tokens ({<think>}, {</think>}) marking the reasoning span. The correct final answer is shown in \textcolor{green}{green}.}}
    \resizebox{0.85\textwidth}{!}{ \begin{tabular}{>{\centering\arraybackslash}m{0.5cm}|>{\raggedright\arraybackslash}m{12.0cm}}
    $\mathbf{x}$ & \texttt{Let $A$, $B$, $C$, and $D$ be point on the hyperbola $\frac{x^2}{20}- \frac{y^2}{24} = 1$ such that $ABCD$ is a rhombus whose diagonals intersect at the origin. Find the greatest real number that is less than $BD^2$ for all such rhombi.} \\
    \midrule
    $\bfr$ & \textbf{\textcolor{red}{<think>}}\texttt{Okay, so I have this problem about a rhombus ABCD on the hyperbola $x^2$/20 - $y^2$/24 = 1. The diagonals intersect at the origin, and I need to find the greatest real number less than BD² for all such rhombi. Hmm, let me start by recalling some properties of hyperbolas and rhombuses.\textbackslash n\textbackslash n First, ...} \textbf{But wait}, \texttt{the problem says the diagonals intersect at the origin. Since the diagonals of a rhombus bisect each other, that means the origin is the midpoint of both diagonals. ...} \textbf{\textcolor{red}{Wait}}, \texttt{BD is from B to D. Coordinates of B: ...} \textbf{\textcolor{red}{Therefore}} \texttt{, the greatest real number less than $BD^2$ for all such rhombi is 480. So, the final answer is \textbackslash boxed\{480\}.\textbackslash n\textbackslash 
 n**Final Answer**\textbackslash n\textbackslash \text{boxed}\{480\}}\textbf{\textcolor{red}{</think>}}\\
    \midrule
        $\mathbf{y}$ & \texttt{..., Thus, the greatest real number less than \textbackslash($BD^2$\textbackslash) for all such rhombi is \textcolor{green}{\textbackslash(\textbackslash boxed\{480\}\textbackslash)}}
    \end{tabular}
}
    \label{tab:aime_example}
\end{table}

\section{Limitations}
\label{appendix: limitation}
While {\ours} presents a practical approach to reducing training costs through thought-level condensation, several limitations remain. First, it relies on a heuristics-based segmentation of CoT traces into head, middle, and tail, which may not align with the true semantic structure of reasoning. Second, the condensation ratio is globally defined and does not adapt to the difficulty of individual examples, more challenging problems may benefit from retaining additional reasoning steps. Third, {\ours} is implemented via supervised fine-tuning and does not explore reinforcement learning (RL)-based training, which could enable more dynamic, reward-driven condensation. Finally, our evaluation is limited to structured mathematical reasoning; extending {\ours} to domains such as open-ended QA or legal reasoning requires further validation.

\section{Broader Impact}
\label{appendix: impact}
This work aims to improve the efficiency of reasoning supervision in large language models by condensing CoT-type reasoning traces. By reducing the length of \textit{training} trajectories, our approach can lower computational costs and carbon footprint, making reasoning-enhanced model training more accessible to researchers and practitioners with limited resources.
However, efficiency gains through condensation may come at the expense of preserving subtle but important reasoning patterns, potentially affecting model robustness, interpretability, or fairness. These potential trade-offs highlight the need for further investigation into how condensation impacts downstream performance across diverse tasks, domains, and user groups. We encourage future work to assess these dimensions and to develop techniques that balance efficiency with reliability and inclusivity.
